\documentclass{article}

\usepackage[preprint, nonatbib]{neurips_2026}

\usepackage[utf8]{inputenc} 
\usepackage[T1]{fontenc}    
\usepackage[table]{xcolor}         
\definecolor{darkorange}{RGB}{180, 45, 0}
\definecolor{PPblue}{RGB}{0, 98, 136}
\definecolor{PPbluelight}{RGB}{0, 126, 164}
\definecolor{PPorange}{RGB}{191, 83, 39}
\definecolor{PPorangeSat}{RGB}{207, 75, 23} 
\definecolor{NRGpurpleSat}{RGB}{81, 10, 194} 
\usepackage[colorlinks=true, allcolors=PPorange]{hyperref}       
\usepackage{url}            
\usepackage{booktabs}       
\usepackage{amsfonts}       
\usepackage{nicefrac}       
\usepackage{microtype}      
\usepackage{graphicx}
\usepackage{xspace}
\usepackage{amsmath}
\usepackage{multirow}
\usepackage{subcaption} 
\usepackage[percent]{overpic}
\usepackage[style=numeric, natbib=true, url=false, sorting=none, maxbibnames=99]{biblatex}
\usepackage{tikz}
\usepackage{textcomp}
\usepackage{siunitx}
\usepackage[cachedir=minted-cache,frozencache]{minted}
\usepackage{diagbox}
\usepackage{xfp}
\definecolor{codebg}{gray}{0.96}
\definecolor{focusbg}{RGB}{255,242,187}
\usetikzlibrary{positioning}
\DeclareFieldFormat[article,inbook,incollection,inproceedings]{title}{#1}
\title{Self-Distilling Generative Adversarial Networks}

\author{%
  Antoni Nowinowski \\
  Poznan University of Technology \\
  \texttt{antoni.nowinowski@cs.put.poznan.pl} \\
  \And
  Krzysztof Krawiec \\
  Poznan University of Technology \\
  \texttt{krzysztof.krawiec@cs.put.poznan.pl}
}

\newcommand{\mname}{SD-GAN\xspace}
\newcommand{\EMA}{\textit{EMA}}
\newcommand{\G}{\ensuremath{G}\xspace}
\newcommand{\Gema}{\ensuremath{G_{\!\EMA}}\xspace}
\newcommand{\thetaEma}{\ensuremath{\theta_{\EMA}}\xspace}
\newcommand{\D}{\ensuremath{D}\xspace}
\newcommand{\Ladv}{\ensuremath{\mathcal{L}_{\textit{adv}}}\xspace}
\newcommand{\Lsd}{\ensuremath{\mathcal{L}_{\textit{SD}}}\xspace}
\newcommand{\thl}{\ensuremath{t_{\nicefrac{1}{2}}}\xspace}
\newcommand{\disc}{\ensuremath{_{\textit{disc}}}\xspace}
\newcommand{\ramp}{\ensuremath{_{\textit{ramp}}}\xspace}
\newcommand{\og}[1]{{\color{black!60}\small(#1)}}

\newcommand{\rl}[2]{
    \pgfmathprintnumber[fixed relative, precision=3]{#2} \\
    \pgfmathprintnumber[fixed relative, precision=3]{#1}
}
\newcommand{\tpm}[2]{
    \num{#1 +- #2}
}
\newcommand{\down}{\,$\boldsymbol{\downarrow}$}
\newcommand{\up}{\,$\boldsymbol{\uparrow}$}
\begin{document}

\maketitle

\begin{abstract}
  In modern GANs, maintaining an Exponential Moving Average (EMA) of the generator's weights is a standard practice, as such an averaged model consistently outperforms the actively trained generator. However, the EMA generator is used for final deployment only and does not influence the training process. To address this missed opportunity, we introduce Self-Distilled GAN (\mname) that employs the EMA generator as a teacher to guide the active generator (student) via perceptual loss.  
  We prove the local asymptotic stability of \mname in the simplified continuous-time Dirac-GAN setting and show that it dampens the parasitic cycling behavior that plagues conventional GANs. Empirical evaluations across established architectures and datasets demonstrate that \mname improves the final image quality on several metrics (FID and random-FID in particular), stabilizes the optimization trajectory and provides additional learning guidance that is not trivially correlated with the conventional adversarial loss. It also proves effective for fine-tuning pretrained GAN models.   
\end{abstract}

\section{Introduction}

\begin{figure}
  \centering\includegraphics[trim={1pt 1pt 1pt 0pt}, clip, width=.7\linewidth]{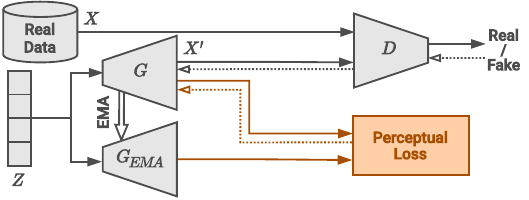}
  \caption{Overview of the \mname framework. The training is guided by both the conventional GAN loss and the self-distillation loss. Dotted lines indicate gradient flow; gradients from the perceptual loss do not propagate through \Gema.}\label{fig:method}
\end{figure}

Obtaining well-performing generators by training generative adversarial networks (GANs) \cite{goodfellow_generative_2014} can be notoriously challenging, one reason being that the game played between the generator (\G) and the discriminator (\D) tends to engage in a cyclic ``chase'' or entirely fail to converge \cite{mescheder_which_2018}. The resulting \G may then fail to capture the entirety of the distribution represented by the training sample or produce samples of inferior quality. To address this challenge, it is common to obtain final generators from GANs using weight averaging,  
which is typically performed outside or inside the training loop. 
In the former case, the averaged model is kept ``on the side'' and does not influence the training process in any way, being used only for the post-training evaluations, as in stochastic weight averaging \cite{izmailov_averaging_2018}.
In the latter case, the averaged model is used as a teacher in self-distillation (SD) student-teacher mode, e.g., in Mean Teachers \cite{tarvainen_mean_2017}, BYOL \cite{grill_bootstrap_2020}, and the DINO family of models \cite{caron_emerging_2021, oquab_dinov2_2024, simeoni_dinov3_2025}. In most of those scenarios, it is common to aggregate the models using exponential moving average (EMA), which naturally attributes more importance to the more recent  models. Specifically, at each training step $t$, the EMA weights \thetaEma are updated using the weights $\theta$ of the active model at decay rate $\beta$:
\begin{equation}
\thetaEma^{(t)} = \beta\theta_{\EMA}^{(t - 1)} + (1 - \beta)\theta^{(t)}.\label{eq:ema}
\end{equation}
In this study, we propose Self-Distilled GAN (\mname) that incorporates SD into GAN training (Fig.\ \ref{fig:method}): rather than learning from the adversarial loss alone, the generator \G also learns from its self-distilled teacher \Gema obtained from \G using EMA. To this end, \G and \Gema are queried on the same latent vector and the similarity of the images they produce is assessed using a popular form of perceptual loss (LPIPS, \cite{zhang_unreasonable_2018}). This changes the training trajectory of \G, and subsequently the discriminator \D. We show that this extension of the traditional GAN paradigm stabilizes training, makes it faster and ultimately results in generators that produce more realistic images. 

\section{Related Work}

\textbf{Weight averaging in GANs}. An early attempt to stabilize GAN training by leveraging the historical states of the players was  \emph{historical averaging} \cite{salimans_improved_2016}, which penalized both \G and \D by the distance between the current and the historical parameters, defined as  $\|\theta^{(t)} - \frac{1}{T}\sum_{i=1}^{T} \theta^{(t-i)}\|^2$. However, it neither maintains a shadow generator for inference nor uses exponential discounting.

To our knowledge, maintaining an EMA of the weights of a GAN generator was first introduced as part of the Progressive Growing of GANs (ProGAN) training framework \cite{karras_progressive_2018}. In order to maintain the generator's responsiveness,
the authors opted not to use learning rate decay. Consequently, the weights fluctuated significantly during training, leading to visual artifacts. To dampen this variance and produce stable outputs, they maintained a passive EMA of the generator's weights. 

The above techniques  were introduced as practical engineering heuristics to improve inference quality, without theoretical justification or citations to the prior literature.
To bridge this gap, \textcite{yazici_unusual_2019} conducted a formal analysis of weight averaging in GANs. They noted that standard gradient descent in adversarial minimax games frequently fails to converge, as the players continuously orbit the optimal solution in limit cycles. 
Through a theoretical analysis, they proved that while EMA does not directly achieve the Nash equilibrium, it exponentially shrinks the amplitude of these limit cycles. Consequently, the active \G model continues exploring the loss landscape, while the passive \Gema is confined to the neighborhood of the equilibrium point. 
Extensive empirical evaluations confirmed that EMA models consistently achieve higher image fidelity than their actively trained counterparts. 
This dynamic has also been studied in \cite{morales-brotons_exponential_2024} across a set of image classification tasks, revealing that EMA models reliably converge to fundamentally different, flatter basins of the loss landscape, not only improving final generalization but yielding remarkable early-stage performance, largely explaining their success when used as teachers.

Following the introduction in ProGAN, EMA of generator weights became a standard component of most state-of-the-art generative models. The technique was adopted in BigGAN \cite{brock_large_2019}, the StyleGAN lineage of models \cite{karras_style-based_2019, karras_analyzing_2020, karras_training_2020, karras_alias-free_2021}, ProjectedGAN \cite{sauer_projected_2021}, StyleGAN-XL \cite{sauer_stylegan-xl_2022}, StyleGAN-T \cite{sauer_stylegan-t_2023}, SAN \cite{takida_san_2024}, and StyleNAT \cite{walton_efficient_2025}, as well as in diffusion models \cite{ho_denoising_2020, song_score-based_2021, dhariwal_diffusion_2021}. However, in contrast to \mname, in all these architectures, \Gema serves as the end-of-run outcome, while not directly impacting the training process.

\textbf{Stabilization and consistency regularization}. Alongside weight averaging, various regularization techniques have been proposed to stabilize the adversarial game. For instance, CR-GAN \cite{zhang_consistency_2020} and bCR \cite{zhao_improved_2021} apply consistency terms to the discriminator, penalizing prediction changes across differently augmented views of the same image, while methods like DiffAugment \cite{zhao_differentiable_2020} apply differentiable augmentations to both real and fake samples to prevent discriminator overfitting. On the generator side, latent consistency regularization (zCR) \cite{zhao_improved_2021} maximizes the difference between images generated from perturbed latent codes to promote diversity. Furthermore, LeCam regularization \cite{tseng_regularizing_2021} utilizes discriminator EMA to anchor its predictions to historical states. While these methods are highly effective, they predominantly prevent discriminator overfitting or enforce latent consistency. In contrast, \mname introduces temporal consistency directly into the generator's trajectory.

\textbf{Knowledge distillation}. In the modern generative landscape, knowledge distillation is widely utilized, though predominantly to improve computational efficiency rather than training stability. This includes distilling lightweight student GANs from larger teacher models \cite{li_revisiting_2021, ren_online_2021}. Concurrently, adversarial distillation techniques have been applied to diffusion models to reduce inference times, enabling high-fidelity sampling in a few steps \cite{sauer_adversarial_2024, sauer_fast_2024} or a single step \cite{kang_distilling_2024}. Beyond model compression, some elements of self-distillation combined with GANs have also been used as a component for a specific supervised downstream task \cite{zheng_semi-tsgan_2024}. In contrast, \mname utilizes it to directly improve the primary model's training trajectory and final image quality. Recently, consistency models \cite{song_consistency_2023}, a new class of diffusion-inspired generative models, have similarly demonstrated the efficacy of EMA-based self-distillation, utilizing a slowly updated target network as a stable reference to guide the active model. Finally, \mname should not be confused with Self-Distilled StyleGAN \cite{mokady_self-distilled_2022}, as the latter focuses on generative self-filtering to curate noisy, multi-modally distributed web data.

\section{Self-Distilled GAN}\label{sec:method}

The motivation for \mname is similar to that postulated in previous methods that engaged weight averaging and self-distillation \cite{tarvainen_mean_2017}: (i) to provide the learner (student) with a form of long-term memory (the teacher) that integrates the effects of the learning process, and (ii) to use the teacher as an anchor to stabilize the training, i.e.\ to avoid catastrophic forgetting of the already acquired knowledge, while allowing for a fair share of exploration. This is particularly relevant in an adversarial game, where \G and \D can keep responding to one another without settling, yielding cyclic parameter trajectories \cite{mescheder_which_2018}. Passive EMA smooths the \Gema used for evaluation but does not affect the update of the active \G. Here, in contrast, \Gema participates in training by providing a smoothed reference that regularizes \G. We implement this idea by extending the traditional GAN adversarial loss function with a self-distillation loss applied to the outputs of the active generator model \G and the averaged model \Gema. Specifically, \G and \Gema are queried on the same random latent vector $z$, and the images they produce are compared using an SD loss component \Lsd. The gradient calculated from \Lsd affects only \G, i.e. \Gema is updated only via EMA  (Fig.\ \ref{fig:method}).

In general, \Lsd can be any image-comparing loss. We use a perceptual loss in the main experiments and examine pixel-wise and discriminator-feature alternatives in Sec.\ \ref{sec:results}. 
Specifically, we utilize LPIPS \cite{zhang_unreasonable_2018}, which evaluates the distance between the feature activations of the two images when passed through a pretrained network. The objective function for the generator is a weighted sum of the standard adversarial loss \Ladv and \Lsd:
\begin{equation}
\mathcal{L}_{G} = \Ladv + \alpha \cdot \Lsd(\G(z), \Gema(z))\label{eq:loss}
\end{equation}
where $\alpha$ controls the weight of the guidance provided by the EMA teacher.
In our implementation, we additionally apply the same differentiable random spatial transformation to both generated images before evaluating \Lsd (Appendix \ref{app:pseudocode}). It is not necessary for the self-distillation itself, but the results of the ablation in Fig.\ \ref{fig:ablations} suggest that this augmentation provides a small additional benefit.

\textbf{Theoretical intuition.}
To better understand the stabilizing mechanism of \mname, we analyze its behavior on the Dirac-GAN \cite{mescheder_which_2018}, a simple, one-dimensional model used to study the convergence of GANs. The Dirac-GAN consists of a true data distribution fixed at zero, a generator modeled as a constant function with a single learnable parameter $G(z) = \theta$, and a linear discriminator $D(x) = \psi \cdot x$. Consequently, the discriminator's score on generated data is $D(G(z)) = \psi\theta$, and the adversarial loss takes the form $\mathcal{L}_{\mathit{adv}} = f(\psi\theta)$ (where $f$ is determined by the GAN variant; here $f$ is identity). The unique Nash equilibrium of this system is $(\theta^*, \psi^*) = (0, 0)$.

\begin{figure}
  \centering\begin{overpic}[trim={7pt 7pt 7pt 7pt}, clip, width=\linewidth]{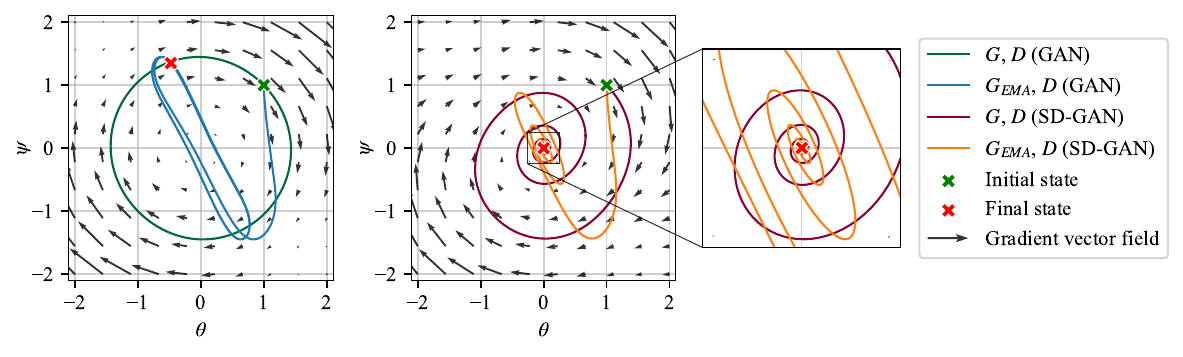}
    \put(0, 2){
      \phantomsubcaption
      \label{fig:Dirac-GAN_a}
      \small (a)
    } 
    \put(29.5, 2){
      \phantomsubcaption
      \label{fig:Dirac-GAN_b}
      \small (b)
    }
  \end{overpic}
  \caption{Parameter trajectories of the active generator ($\theta$), EMA generator ($\theta_{\textit{EMA}}$) and discriminator ($\psi$) in the Dirac-GAN. \textbf{(a)} Standard GAN forms cycles orbiting the Nash equilibrium. \textbf{(b)} Self-distillation dampens these cycles into a contracting spiral. The arrows visualize the gradient vector field (b) for the final $\theta_{\textit{EMA}}$.}\label{fig:Dirac-GAN}
\end{figure}

Figure \ref{fig:Dirac-GAN} illustrates the trajectories of these parameters during training. In the basic GAN (\ref{fig:Dirac-GAN_a}, green plot), gradient descent does not converge, with $G$ and $D$ persistently orbiting the equilibrium; incorporating EMA (blue plot) narrows the orbit, but does not improve convergence.
The right inset (\ref{fig:Dirac-GAN_b}) demonstrates how the training dynamics change with the addition of our self-distillation objective \Lsd (simplified to $\alpha( \G(z) - \Gema(z))^2$). By continuously penalizing the distance between the active generator and the EMA generator, \Lsd introduces a damping force that transforms the circular, non-converging trajectory into a contracting spiral, forcing the parameters to converge in the limit to the exact Nash equilibrium. This happens both for the EMA (orange) and the non-EMA (indigo) configuration. This analytical model corroborates our empirical findings in the simplified Dirac-GAN scenario: by anchoring the active model to its historical average, \mname mitigates oscillatory training dynamics and ensures stable convergence.


\textbf{Proof of convergence.}
We follow the above intuition with a more rigorous formalization by examining the gradient vector field and deriving the exact local convergence criterion. As in prior analyses of local GAN stability, we study the continuous-time dynamics through the spectrum of the Jacobian of the vector field at the equilibrium \cite{khalil_nonlinear_1996, mescheder_numerics_2017, nagarajan_gradient_2017, mescheder_which_2018}. We consider a continuous-time formulation of the Dirac-GAN minimax game. We denote the active generator parameter as $\theta(t)$, the discriminator parameter as $\psi(t)$, and the EMA generator parameter as $\phi(t)$ (rather than $\theta_{\textit{EMA}}\!(t)$, for brevity), and introduce continuous-time learning rates $\eta_G, \eta_D > 0$ for the generator and discriminator respectively. For the adversarial objective $\mathcal{L}_{\mathit{adv}}(\theta,\psi) = f(\psi\theta)$, the local behavior near the Nash equilibrium depends only on $c = f'(0)$, with $c \neq 0$, so we work with the linearized loss function $\mathcal{L}_{\mathit{adv}} = c\psi\theta$. The self-distillation penalty is $\alpha \cdot \mathcal{L}_{SD} = \tfrac{\alpha}{2} (\theta - \phi)^2$ (the $\tfrac{1}{2}$ factor is for convenience). We parametrize the EMA decay rate $\beta$ as an update rate $\eta_\phi = 1 - \beta$. The joint vector field is therefore:
\begin{equation}
\left\{
\begin{aligned}
\tfrac{d\theta}{dt} &= -\eta_G (c \psi + \alpha(\theta - \phi)) \\
\tfrac{d\psi}{dt} &= \eta_D c \theta \\
\tfrac{d\phi}{dt} &= \eta_\phi (\theta - \phi)
\end{aligned}
\right.
\end{equation}
To analyze local stability around the Nash equilibrium $(\theta^*, \psi^*, \phi^*) = (0,0,0)$, we construct the Jacobian matrix $J$ of the state vector $(\theta, \psi, \phi)$:
\begin{equation}
J = \begin{bmatrix}
-\eta_G \alpha & -\eta_G c & \eta_G \alpha \\
\eta_D c & 0 & 0 \\
\eta_\phi & 0 & -\eta_\phi
\end{bmatrix}
\end{equation}
The system converges asymptotically to the Nash equilibrium if and only if all eigenvalues $\lambda$ of $J$ have negative real parts \cite{khalil_nonlinear_1996}. We solve the characteristic equation $\det(\lambda I - J) = 0$ by expanding along the second row:
\begin{gather}
\lambda^3 + (\eta_G\alpha + \eta_\phi)\lambda^2 + (\eta_D \eta_G c^2)\lambda + \eta_D \eta_G \eta_\phi c^2 = 0
\end{gather}
To make sure the eigenvalues $\lambda$ have negative real parts, we apply the Routh--Hurwitz criterion \cite{khalil_nonlinear_1996}, which states that all roots of a 3rd-degree polynomial ($a_3\lambda^3 + a_2\lambda^2 + a_1\lambda + a_0 = 0$) will have strictly negative real parts if and only if (i) $a_3, a_2, a_1, a_0 > 0$ and  (ii) $a_2 a_1 > a_3 a_0$:
\begin{equation}
\text{(i)}\quad
\begin{alignedat}{2}
a_3{:}&\ 1 > 0 \qquad & a_2{:}&\ \eta_G \alpha + \eta_\phi > 0 \\
a_1{:}&\ \eta_D \eta_G c^2 > 0 \qquad & a_0{:}&\ \eta_D \eta_G \eta_\phi c^2 > 0
\end{alignedat}
\qquad \text{(ii)}\quad \eta_D \eta_G^2 c^2 \alpha > 0
\end{equation}
Because the learning rates $\eta_G, \eta_D > 0$, the condition simplifies to $\alpha > 0$. Therefore, for any valid learning rates $\eta_G, \eta_D > 0$, and EMA decay rate $\beta \in [0, 1)$, introducing a self-distillation penalty $\alpha > 0$ strictly guarantees that the eigenvalues have negative real parts, proving local asymptotic convergence to the exact Nash equilibrium, rather than the oscillatory behavior of the unregularized Dirac-GAN \cite{mescheder_which_2018}.

\section{Results}\label{sec:results}

We validate \mname by applying it to the well-established GAN architectures StyleGAN2-ADA \cite{karras_training_2020}, ProjectedGAN \cite{sauer_projected_2021} and StyleNAT \cite{walton_efficient_2025}, and experimenting with the widely used benchmarks FFHQ \cite{karras_style-based_2019} and LSUN Church \cite{yu_lsun_2016} at $256^2$ resolution. We additionally evaluate StyleGAN-XL \cite{sauer_stylegan-xl_2022} on ImageNet at $64^2$ resolution. We take care to faithfully reproduce the architectures and results of the cited works by relying on the original implementations (including metric calculation), and amending them only as minimally as required to extend them to \mname. See Appendix \ref{app:soft-impl} for further  details. Despite our best efforts, we did not manage to replicate the results of StyleNAT on LSUN Church sufficiently well, so we discard this combination (see Appendix\ \ref{app:replication} for details). In addition to the default version of \mname that uses LPIPS as \Lsd, we also consider a simplified variant that uses pixel-wise $L_1$ for that purpose. We set $\alpha$ to $1$ for StyleGAN2-ADA and ProjectedGAN, $0.25$ for StyleNAT, and $100$ for StyleGAN-XL (the latter two adjusted for their \Ladv scales). The values of $\beta$ are set to the default values of the baselines to maintain a fair comparison.

Similar to other works, we use the Fréchet Inception Distance (FID, \cite{heusel_gans_2017}) as the primary metric for measuring the quality of obtained generators. The FID estimates are calculated from a sample of $50\text{k}$ generated images and compared against the full dataset. However, to address the growing criticism regarding its vulnerability to ImageNet-specific biases \cite{sauer_stylegan-xl_2022, stein_exposing_2023}, we complement it with the random-FID (rFID, \cite{sauer_stylegan-xl_2022}). By employing an untrained feature extractor, rFID helps verify if improvements on FID result from fundamental gains in image quality rather than metric-specific artifacts. 
Furthermore, to rigorously test whether the self-distillation objective induces mode collapse or reduces diversity, we use Precision and Recall \cite{sajjadi_assessing_2018}. Precision estimates how much of the generated distribution lies near real examples, while Recall estimates how much of the real distribution is covered by generated examples. All reported results concern the EMA models. 

\begin{figure}
  \centering\includegraphics[trim={7pt 7pt 7pt 7pt}, clip, width=\linewidth]{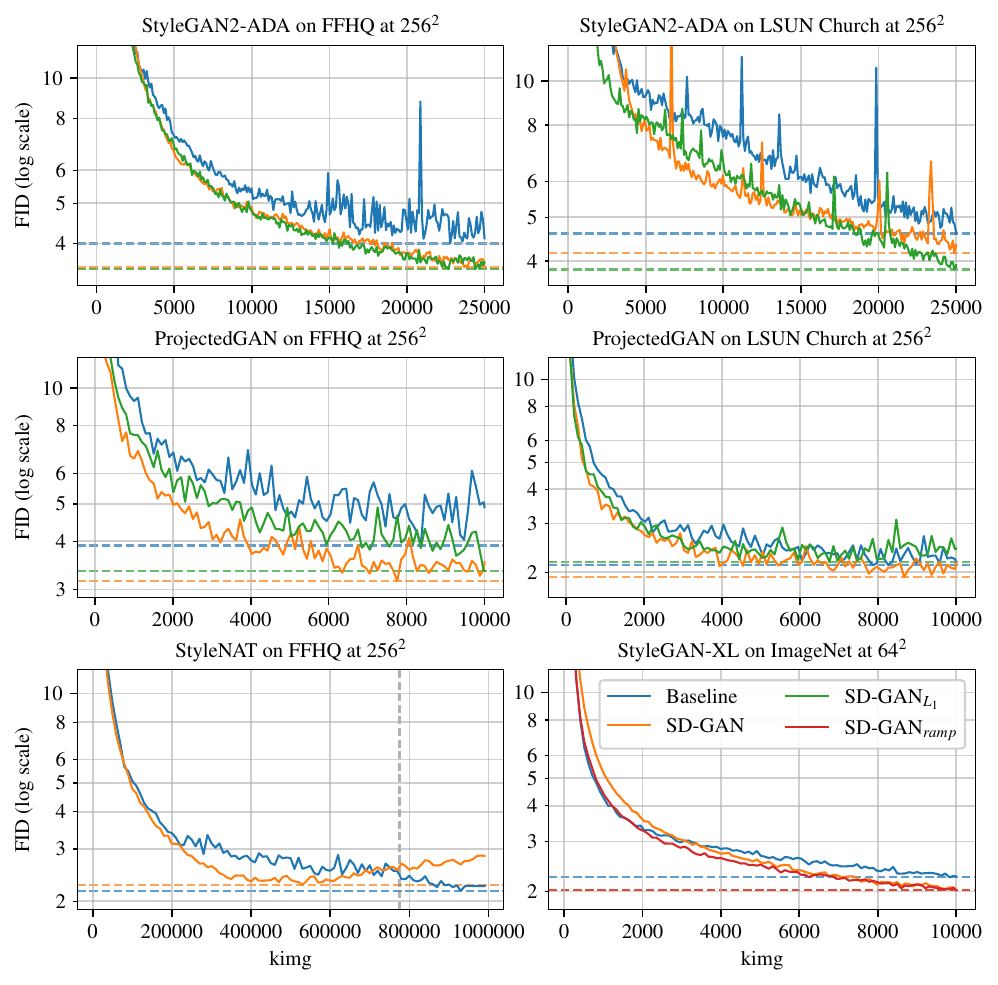}
  \caption{FID scores (log scale) during training on the FFHQ and LSUN Church datasets across three baseline architectures. Following \cite{walton_efficient_2025}, we allowed StyleNAT to learn much longer than the remaining architectures. Horizontal dashed lines indicate the best FID score. 
  }\label{fig:training}
\end{figure}

Figure \ref{fig:training} presents the FID over training effort expressed in thousands of processed real images (kimgs). \mname systematically achieves better FID values (notice log scale), and for the convolutional architectures (StyleGAN2-ADA and ProjectedGAN) does so already from the early iterations of learning. For ProjectedGAN, \mname using LPIPS achieves the same FID as the baseline method at a fraction of the cost ($\sim32\%$ on FFHQ and $\sim39\%$ on LSUN Church). Using $L_1$ as \Lsd leads to somewhat worse results than LPIPS, however, usually still better than the baseline. The smoother curves for \mname suggest that rather than slowing  exploration or anchoring to the historical state, the distillation objective acts as a stabilizing force, corroborating the theoretical findings from Sec.\ \ref{sec:method}.  

Table \ref{tab:FID_comparison} compares the final metrics achieved by reference methods and their \mname variants. As we observed rFID to vary significantly across independent evaluations due to random feature projection inherent to this metric \cite{sauer_stylegan-xl_2022}, we compute it five times and report means and standard deviations. The advantage of self-distillation is particularly prominent on this metric, especially for the LPIPS-based \mname; e.g., the ProjectedGAN's rFID drops from \tpm{79.344}{7.644} to \tpm{14.103}{2.091} on FFHQ. Thus, the EMA teacher gains hold also for an untrained feature extractor (beyond ImageNet-biased FID). When applied to StyleGAN2-ADA and ProjectedGAN, \mname also tends to improve Recall or Precision, while only moderately deteriorating the other metric, demonstrating that the model maintains diversity and, contrary to potential concerns, is not overly anchored to the averaged state by self-distillation. While \mname yields consistent benefits for convolutional baselines, we observe a slight FID regression for the transformer-based StyleNAT (2.17 to 2.27 on FFHQ), despite initial gains visible in Fig.\ \ref{fig:training}, although the baseline overtook the SD variant only after learning-rate decay began (vertical dashed line). 
For StyleGAN-XL+SD\ramp, we delay SD and ramp up $\alpha$ as discriminator blurring \cite{sauer_stylegan-xl_2022,karras_alias-free_2021} fades, to avoid reinforcing high-frequency details hidden by blur.



\begin{table}[t!]
\caption{Comparison of evaluation metrics (FID, rFID, Precision, and Recall) on FFHQ and LSUN Church at $256^2$ and ImageNet at $64^2$. Values in parentheses indicate the original reported scores from the respective papers (see Appendix \ref{app:replication} for explanations of the discrepancies). }
\label{tab:FID_comparison}
\centering
\begin{tabular}{llllll}
\toprule
Dataset & Model & FID\down & rFID\down & Precision\up & Recall\up \\
\midrule
\multirow{8}{*}[-1.5ex]{\begin{tabular}[c]{@{}l@{}}FFHQ\\[0.2ex]($256^2$)\end{tabular}}
 & StyleGAN2-ADA    & 4.00 \og{4.30} & \tpm{1.584}{0.112}  & 0.692 & 0.399 \\
 & StyleGAN2-ADA+SD & 3.50           & \tpm{1.181}{0.274}  & 0.678 & 0.440 \\
 & StyleGAN2-ADA+SD$_{L_1}$ & 3.46      & \tpm{4.509}{0.904}  & 0.679 & 0.469 \\
\addlinespace
 & ProjectedGAN     & 3.90 \og{3.39} & \tpm{79.344}{7.644} & 0.662 & 0.439 \\
 & ProjectedGAN+SD  & 3.15           & \tpm{14.103}{2.091} & 0.666 & 0.452 \\
 & ProjectedGAN+SD$_{L_1}$ & 3.35    & \tpm{34.394}{3.555} & 0.669 & 0.446 \\
\addlinespace
 & StyleNAT         & 2.17 \og{2.05} & \tpm{6.307}{0.554}  & 0.698 & 0.383 \\
 & StyleNAT+SD      & 2.27           & \tpm{4.223}{0.699}  & 0.676 & 0.367 \\
\midrule
\multirow{6}{*}[-0.5ex]{\begin{tabular}[c]{@{}l@{}}LSUN\\Church\\[0.2ex]($256^2$)\end{tabular}}
 & StyleGAN2-ADA    & 4.60           & \tpm{2.727}{0.184}  & 0.626 & 0.218 \\
 & StyleGAN2-ADA+SD & 4.16           & \tpm{1.807}{0.198}  & 0.619 & 0.274 \\
 & StyleGAN2-ADA+SD$_{L_1}$ & 3.83      & \tpm{0.977}{0.075}  & 0.622 & 0.321 \\
\addlinespace
 & ProjectedGAN     & 2.11 \og{1.59} & \tpm{5.725}{1.256}  & 0.561 & 0.474 \\
 & ProjectedGAN+SD  & 1.91           & \tpm{1.917}{0.279}  & 0.605 & 0.446 \\
 & ProjectedGAN+SD$_{L_1}$ & 2.18    & \tpm{2.950}{0.462}  & 0.607 & 0.429 \\
\midrule
\multirow{3}{*}{\begin{tabular}[c]{@{}l@{}}ImageNet\\[0.2ex]($64^2$)\end{tabular}}
 & StyleGAN-XL         & 2.24 \og{1.51} & \tpm{1.114}{0.284}  & 0.690 & 0.507 \\
 & StyleGAN-XL+SD      & 2.02           & \tpm{1.782}{0.945}  & 0.689 & 0.510 \\
 & StyleGAN-XL+SD\ramp & 2.02           & \tpm{1.543}{0.620}  & 0.697 & 0.508 \\
\bottomrule
\end{tabular}
\end{table}


Table \ref{tab:computational_cost} presents the training times and relative computational overheads of training \mname on a single GPU. 
See Appendix \ref{app:soft-impl} for the details on our computational infrastructure. 

\begin{table}[t!]
\centering
\caption{Computational requirements of \mname training compared to the baseline methods: the time (in seconds) required to process 1 kimg and the average memory consumption. }
\label{tab:computational_cost}
\begin{tabular}{lcccccc}
\toprule
\multirow{2}{*}[-0.5ex]{Method} & \multicolumn{2}{c}{StyleGAN2-ADA} & \multicolumn{2}{c}{ProjectedGAN} & \multicolumn{2}{c}{StyleNAT} \\
\cmidrule(lr){2-3} \cmidrule(lr){4-5} \cmidrule(lr){6-7}
 & s/kimg & Mem. & s/kimg & Mem. & s/kimg & Mem. \\
\midrule
Baseline & 12.56 s & 3.10 GB & 18.78 s & 7.45 GB & 213 s & 29.7 GB \\
\midrule
\mname{$_{L_1}$}   & 14.09 s & 3.36 GB & 22.20 s & 8.81 GB &  &  \\
\mname   & 19.93 s & 5.00 GB & 28.33 s & 9.38 GB & 233 s & 31.8 GB \\
\midrule
\mname{$_{L_1}$} overheads & 
+\fpeval{round((14.09-12.56)/12.56*100, 1)}\% & 
+\fpeval{round((3.36-3.10)/3.10*100, 1)}\% & 
+\fpeval{round((22.20-18.78)/18.78*100, 1)}\% & 
+\fpeval{round((8.81-7.45)/7.45*100, 1)}\% & 
 & 
 \\
\mname overheads & 
+\fpeval{round((19.93-12.56)/12.56*100, 1)}\% & 
+\fpeval{round((5.00-3.10)/3.10*100, 1)}\% & 
+\fpeval{round((28.33-18.78)/18.78*100, 1)}\% & 
+\fpeval{round((9.38-7.45)/7.45*100, 1)}\% & 
+\fpeval{round((233-213)/213*100, 1)}\% & 
+\fpeval{round((31.8-29.7)/29.7*100, 1)}\% \\
\bottomrule
\end{tabular}
\end{table}

\subsection{Sensitivity to hyperparameter settings}\label{app:sensitivity}

We conducted sensitivity analysis for ProjectedGAN on FFHQ at $256^2$ across a grid of self-distillation weights ($\alpha \in \{0.5, 1.0, 2.0\}$) and EMA decay rates $\beta$, set as to correspond to half-life ($\thl \in \{10, 20, 40\}\ \text{kimg}$), so the decay rate is defined as $\beta=0.5^{b / (1000\thl)}$
where batch size $b=64$. Table \ref{tab:sensitivity} summarizes the resulting FID scores. The default configuration ($\alpha = 1.0$, $\thl = 20\ \text{kimg}$, $\text{FID} = 3.15$) used in the experiments reported in Table \ref{tab:FID_comparison} corresponds to the center cell ($\text{FID} = 3.21$, the difference from Table \ref{tab:FID_comparison} reflects normal run-to-run variance). 

Notably, each of these hyperparameter combinations outperforms the baseline ProjectedGAN without SD (FID of 3.90, cf. Table \ref{tab:FID_comparison}), demonstrating that the benefit of EMA-based self-distillation is robust and not overly sensitive to precise tuning. Doubling both the EMA half-life (20 to 40) and the SD loss weight (1 to 2) yields significant gains, reducing the FID score down to 2.74. While this grid search is limited in scope and density of sampling, the manifesting monotonic improvement of FID w.r.t. both considered hyperparameters signals a stable dependency that is likely to hold for other values of hyperparameters.

\begin{table}[t!]
\caption{FID scores for combinations of hyperparameters, ProjectedGAN+SD, FFHQ at $256^2$. }\label{tab:sensitivity}
\centering
\begin{tabular}{lccc}
\toprule
$\alpha$ \hspace{1cm} \thl: & 10 & 20 & 40 \\ \midrule
0.5 & {\cellcolor[HTML]{FBC8C5}} \color[HTML]{000000} 3.57 & {\cellcolor[HTML]{FCD4C4}} \color[HTML]{000000} 3.48 & {\cellcolor[HTML]{FCD9C3}} \color[HTML]{000000} 3.44 \\
1.0 & {\cellcolor[HTML]{FBC8C5}} \color[HTML]{000000} 3.57 & {\cellcolor[HTML]{FEF8C0}} \color[HTML]{000000} 3.21 & {\cellcolor[HTML]{EDF8C8}} \color[HTML]{000000} 2.96 \\
2.0 & {\cellcolor[HTML]{FFFDBF}} \color[HTML]{000000} 3.17 & {\cellcolor[HTML]{F4FAC5}} \color[HTML]{000000} 3.03 & {\cellcolor[HTML]{D9F0D3}} \color[HTML]{000000} 2.74 \\
\bottomrule
\end{tabular}
\end{table}

\subsection{\mname as a fine-tuning technique}

\begin{figure}
  \centering\begin{overpic}
  [trim={7pt 4pt 7pt 6pt}, clip, width=\linewidth]{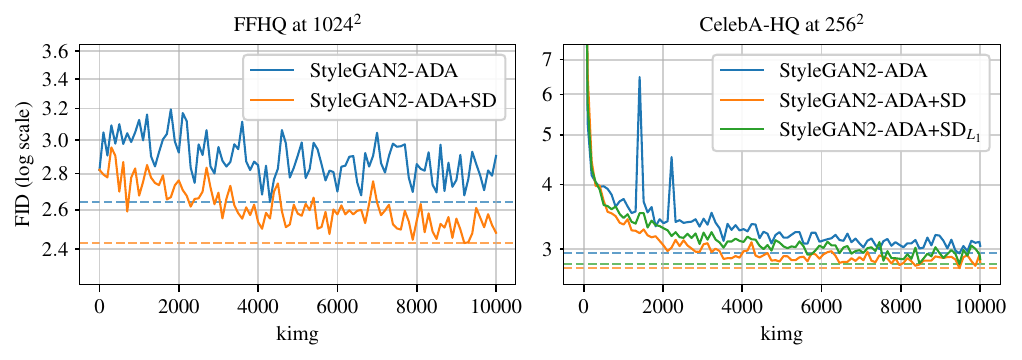}
  \put(4.5, 1.5){
    \phantomsubcaption
    \label{fig:fine-tuning_a}
    \small (a)
  }
  \put(53, 1.5){
    \phantomsubcaption
    \label{fig:fine-tuning_b}
    \small (b)
  }
  \end{overpic}
  \caption{FID scores during fine-tuning on the FFHQ at $1024^2$ and CelebA-HQ at $256^2$.}\label{fig:fine-tuning}
\end{figure}

Motivated by the significant computational overheads reported in Table \ref{tab:computational_cost}, we attempt to engage \mname as a fine-tuning technique. Given a pretrained GAN model, we initialize \G, \Gema and \D with the pretrained weights and continue training using a combination of the standard adversarial loss (\Ladv) and our self-distillation loss (\Lsd). Within each comparison, the baseline and SD variants start from the same pretrained checkpoint and retain their native resolution. Figure \ref{fig:fine-tuning} juxtaposes this with the baseline scenario, which continues training with \Ladv only. 
In Fig.\ \ref{fig:fine-tuning_a} we fine-tune the StyleGAN2-ADA\footnote{Technically we use the StyleGAN2 checkpoints, as recommended in the official repository. \label{fn:SG2}} authors' official checkpoint of the model trained on FFHQ at $1024^2$ resolution \cite{karras_training_2020}, and subject it to training on the same dataset. \mname quickly improves on FID in this megapixel-scale fine-tuning run, while the baseline does not show clear signs of progress. 
In Fig.~\ref{fig:fine-tuning_b} we examine a more realistic fine-tuning scenario, starting from the FFHQ at $256^2$ checkpoint\footref{fn:SG2} and fine-tuning it on the train split of CelebA-HQ dataset at $256^2$ \cite{liu_deep_2015, karras_progressive_2018}. The change of training distribution causes all three models to start from moderate FID values, but they rapidly adapt. Both SD variants not only stabilize the fine-tuning process but also converge to a lower final FID.


\subsection{Ablations}

\begin{figure}
  \centering\includegraphics[trim={7pt 7pt 7pt 7pt}, clip, width=\linewidth]{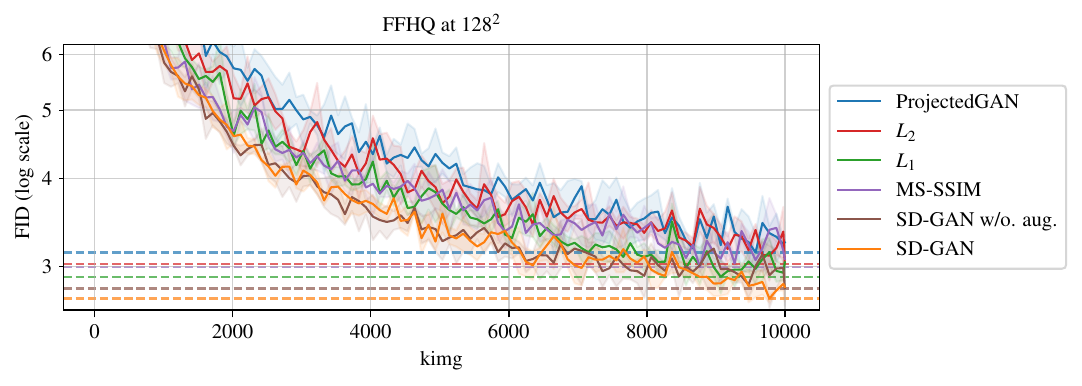}
  \caption{Learning curves for \mname and its ablated variants.}\label{fig:ablations}
\end{figure}

Figure \ref{fig:ablations} compares, for the ProjectedGAN applied to FFHQ at $128^2$, the learning dynamics of the complete \mname to its ablations obtained by disabling augmentations (\mname \emph{w/o. aug.}) or replacing the LPIPS-based \Lsd with simpler distillation losses, which directly compare the images produced by \G and \Gema: $L_1$, $L_2$, and Multi-Scale Structural Similarity Index Measure (MS-SSIM, \cite{1292216}). Combining the latter variants with augmentations would be pointless, so we do not engage them. The figure presents the averages of three runs with shaded standard deviations. While the FID values still fluctuate quite significantly, the LPIPS variants quite clearly converge at the fastest pace, and the augmentations have slight positive impact. 
A separate comparison with delayed-checkpoint weight-space regularization is reported in Appendix \ref{app:temporal-baselines}.

\subsection{SD-GAN with discriminator-based self-distillation loss}\label{sec:disc-based}

By being pre-trained on a huge volume of images, LPIPS-based \Lsd may suffer from ``feature pollution/leakage''. To address this, we consider a \mname variant that calculates \Lsd using the discriminator. To this aim, we query $D$ on the outputs of \G and \Gema (Eq.\ \ref{eq:ema}), and extract the intermediate activations from the first three blocks of both queries. Following the LPIPS methodology, we apply channel normalization to these feature maps and compute their $L_1$ distance. We set $\alpha=10$ to ensure similar contribution to $\mathcal{L}_{G}$ as in the LPIPS variant. 

\begin{table}
\caption{Comparison of evaluation metrics (FID, rFID, Precision, and Recall) on FFHQ at $256^2$. }
\label{tab:FID_comparison_disc}
\centering
\begin{tabular}{lllll}
\toprule
Model                 & FID\down & rFID\down & Precision\up & Recall\up \\
\midrule
StyleGAN2-ADA         & 4.00 & \tpm{1.584}{0.112}  & 0.692 & 0.399 \\
StyleGAN2-ADA+SD      & 3.50 & \tpm{1.181}{0.274}  & 0.678 & 0.440 \\
StyleGAN2-ADA+SD\disc & 3.52 & \tpm{1.589}{0.246}  & 0.682 & 0.446 \\
\bottomrule
\end{tabular}
\end{table}

As shown in Table \ref{tab:FID_comparison_disc}, using the discriminator's own feature space for self-distillation works remarkably well. The SD\disc configuration achieves FID of 3.52, close to the LPIPS-based variant (3.50, Table~\ref{tab:FID_comparison}), without an additional pretrained perceptual network. Its rFID (1.589) is essentially equal to the baseline (1.584) and higher than that of the LPIPS variant (1.181).

\subsection{The interplay between loss components}

\begin{figure}
    \centering
    \begin{subfigure}[b]{\textwidth}
    \centering
    \begin{tikzpicture}[
        node distance = 0.5pt,
        every node/.style = {inner sep=0pt}
    ]
    
      \node (tl1) {\includegraphics[width=0.12\linewidth]{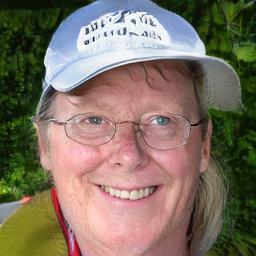}};

      \node (tl2) [right=of tl1] {\includegraphics[width=0.12\linewidth]{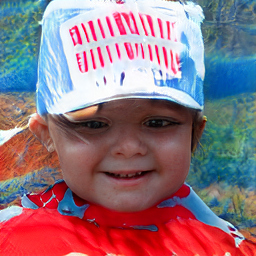}};

      \node (tl3) [right=of tl2] {\includegraphics[width=0.12\linewidth]{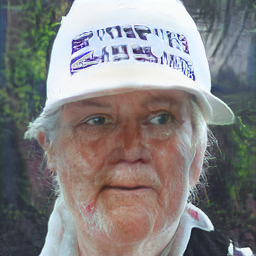}};

      \node (tl4) [right=of tl3] {\includegraphics[width=0.12\linewidth]{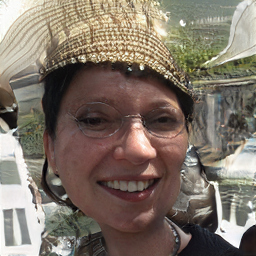}};

      \node (tr1) [right=14pt of tl4] {\includegraphics[width=0.12\linewidth]{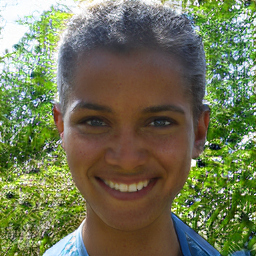}};

      \node (tr2) [right=of tr1] {\includegraphics[width=0.12\linewidth]{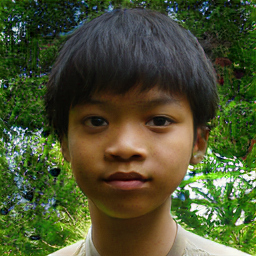}};

      \node (tr3) [right=of tr2] {\includegraphics[width=0.12\linewidth]{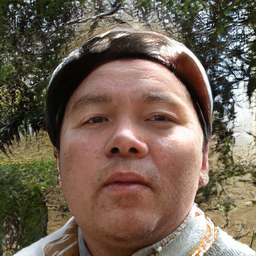}};

      \node (tr4) [right=of tr3] {\includegraphics[width=0.12\linewidth]{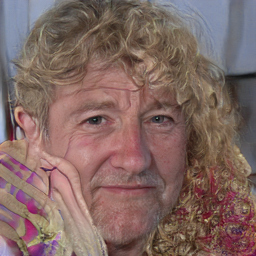}};

      \node (bl1) [below=14pt of tl1] {\includegraphics[width=0.12\linewidth]{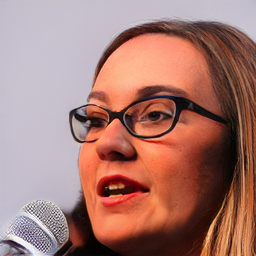}};

      \node (bl2) [right=of bl1] {\includegraphics[width=0.12\linewidth]{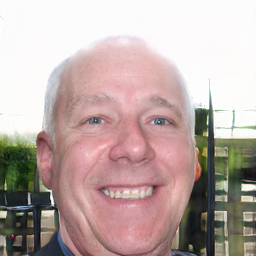}};

      \node (bl3) [right=of bl2] {\includegraphics[width=0.12\linewidth]{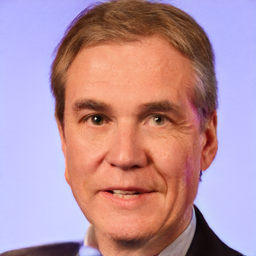}};

      \node (bl4) [right=of bl3] {\includegraphics[width=0.12\linewidth]{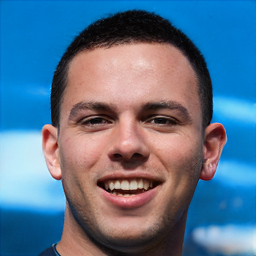}};

      \node (br1) [right=14pt of bl4] {\includegraphics[width=0.12\linewidth]{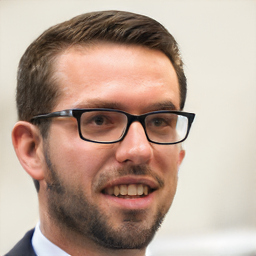}};

      \node (br2) [right=of br1] {\includegraphics[width=0.12\linewidth]{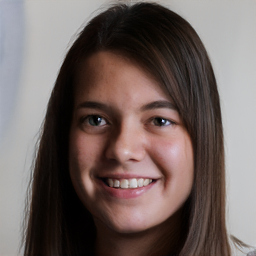}};

      \node (br3) [right=of br2] {\includegraphics[width=0.12\linewidth]{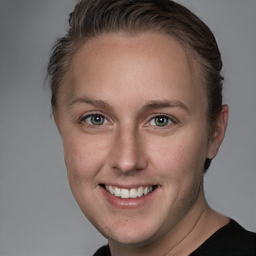}};

      \node (br4) [right=of br3] {\includegraphics[width=0.12\linewidth]{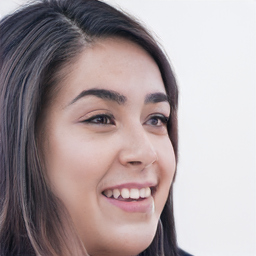}};

      \path (tl4.east) -- (tr1.west) coordinate[midway] (x-mid);
      \path (tl1.south) -- (bl1.north) coordinate[midway] (y-mid);
      \draw[<->, thick] (tl1.west |- y-mid) -- (tr4.east |- y-mid);
      \draw[<->, thick] (x-mid |- bl1.south) -- (x-mid |- tl1.north);
      \node [below=4pt of tl1.south, font=\footnotesize, align=right, fill=white] {\xspace Min $D(x)$ \xspace};
      \node [below=4pt of tr4.south, font=\footnotesize, align=right, fill=white] {\xspace Max $D(x)$ \xspace};
      \node [right=4pt of tl4.east, font=\footnotesize, align=center, fill=white] {\rotatebox{90}{\xspace Worst \Lsd \xspace}}; 
      \node [right=4pt of bl4.east, font=\footnotesize, align=center, fill=white] {\rotatebox{90}{\xspace Best \Lsd \xspace}};

    \end{tikzpicture}
    \caption{StyleGAN2-ADA+SD, FFHQ at $256^2$.}
    \vspace*{2mm}
    \end{subfigure}

    \begin{subfigure}[b]{\textwidth}
    \centering
    \begin{tikzpicture}[
        node distance = 0.5pt,
        every node/.style = {inner sep=0pt}
    ]
    
      \node (tl1) {\includegraphics[width=0.12\linewidth]{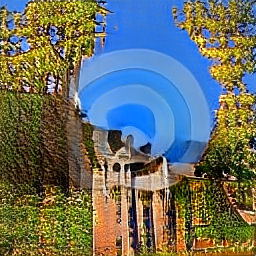}};

      \node (tl2) [right=of tl1] {\includegraphics[width=0.12\linewidth]{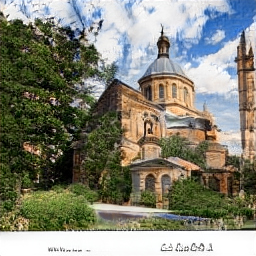}};

      \node (tl3) [right=of tl2] {\includegraphics[width=0.12\linewidth]{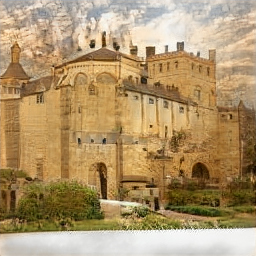}};

      \node (tl4) [right=of tl3] {\includegraphics[width=0.12\linewidth]{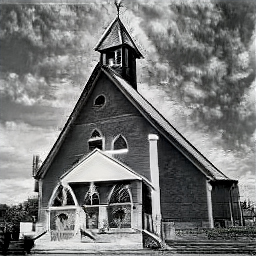}};

      \node (tr1) [right=14pt of tl4] {\includegraphics[width=0.12\linewidth]{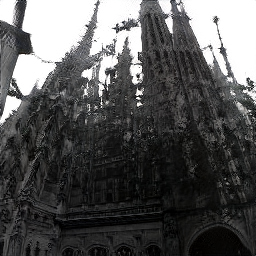}};

      \node (tr2) [right=of tr1] {\includegraphics[width=0.12\linewidth]{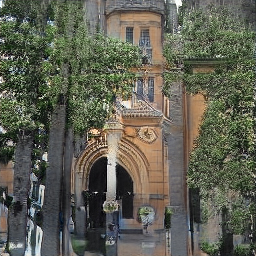}};

      \node (tr3) [right=of tr2] {\includegraphics[width=0.12\linewidth]{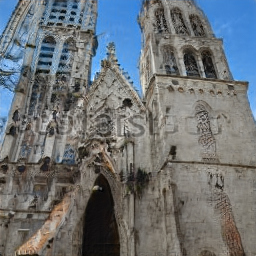}};

      \node (tr4) [right=of tr3] {\includegraphics[width=0.12\linewidth]{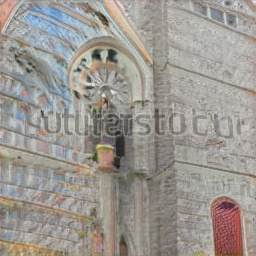}};

      \node (bl1) [below=14pt of tl1] {\includegraphics[width=0.12\linewidth]{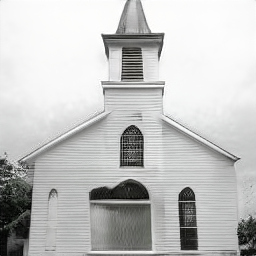}};

      \node (bl2) [right=of bl1] {\includegraphics[width=0.12\linewidth]{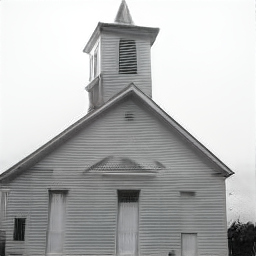}};

      \node (bl3) [right=of bl2] {\includegraphics[width=0.12\linewidth]{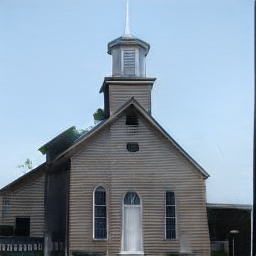}};

      \node (bl4) [right=of bl3] {\includegraphics[width=0.12\linewidth]{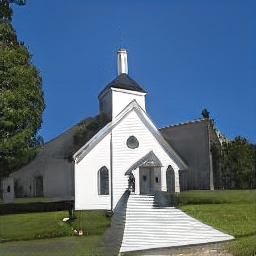}};

      \node (br1) [right=14pt of bl4] {\includegraphics[width=0.12\linewidth]{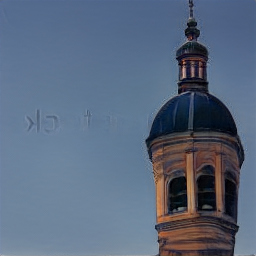}};

      \node (br2) [right=of br1] {\includegraphics[width=0.12\linewidth]{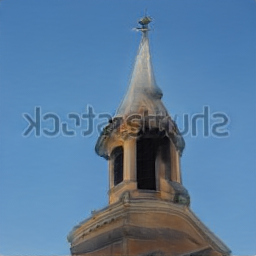}};

      \node (br3) [right=of br2] {\includegraphics[width=0.12\linewidth]{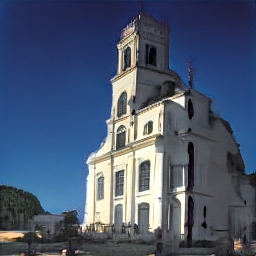}};

      \node (br4) [right=of br3] {\includegraphics[width=0.12\linewidth]{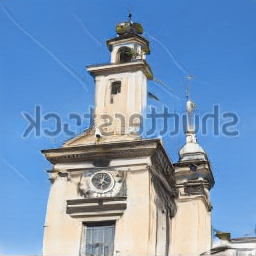}};

      \path (tl4.east) -- (tr1.west) coordinate[midway] (x-mid);
      \path (tl1.south) -- (bl1.north) coordinate[midway] (y-mid);
      \draw[<->, thick] (tl1.west |- y-mid) -- (tr4.east |- y-mid);
      \draw[<->, thick] (x-mid |- bl1.south) -- (x-mid |- tl1.north);

      \node [below=4pt of tl1.south, font=\footnotesize, align=right, fill=white] {\xspace Min $D(x)$ \xspace};
      \node [below=4pt of tr4.south, font=\footnotesize, align=right, fill=white] {\xspace Max $D(x)$ \xspace};
      \node [right=4pt of tl4.east, font=\footnotesize, align=center, fill=white] {\rotatebox{90}{\xspace Worst \Lsd \xspace}}; 
      \node [right=4pt of bl4.east, font=\footnotesize, align=center, fill=white] {\rotatebox{90}{\xspace Best \Lsd \xspace}}; 
    \end{tikzpicture}
    \caption{StyleGAN2-ADA+SD, LSUN Church at $256^2$.}
    \end{subfigure}
    
    \caption{Qualitative interplay between the objectives. \textbf{Left vs. right:} smaller vs. larger \textit{realness} according to \D. \textbf{Top vs. bottom:} larger vs. smaller difference between \G and \Gema (smaller \Lsd).  The adversarial objective pulls generator to the right, while the self-distillation objective pulls it to the bottom.  See Fig.\ \ref{fig:sd_vs_adv_detailed} for per-image values of \Lsd and $D$. }
    \label{fig:sd_vs_adv}
\end{figure}
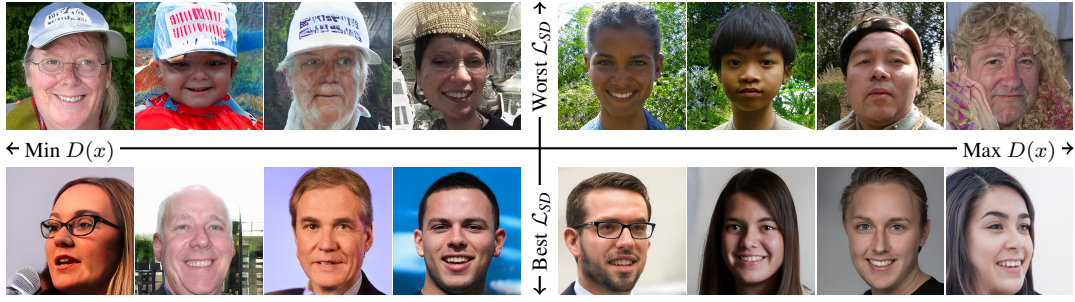
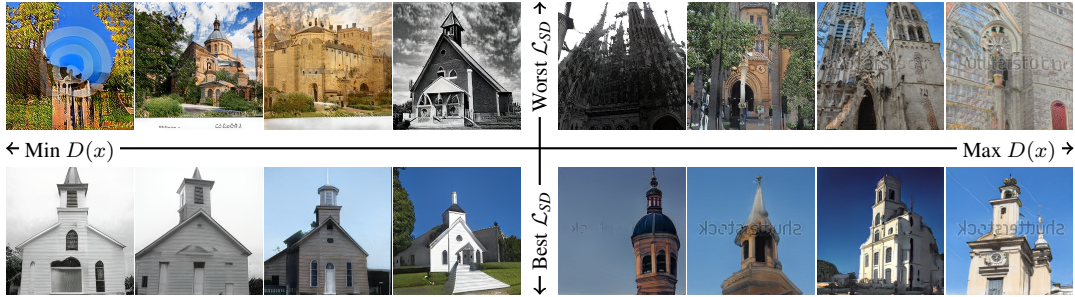

To better understand the interplay between the adversarial and self-distillation objectives, we analyze the visual characteristics of the images favored by each loss component. In Fig. \ref{fig:sd_vs_adv}, we juxtapose selected images obtained by querying the final \Gema models on a sample of $10\text{k}$ latent vectors. For each latent, we generate images using both \Gema and the active generator \G, compute the LPIPS distance \Lsd between them, and select the $200$ images with the largest and smallest \Lsd (respectively top and bottom row in the figure). Then we query the discriminator on the \Gema outputs and select the $4$ highest and lowest scoring images, placing them respectively in the right and left quadrants. The figure presents thus the extremes of the joint distribution of \Lsd (vertical) and \Ladv (horizontal), rather than a representative sample of the generated distribution, and illustrates the distinct image characteristics associated with the two scores. Specifically, the images with worse \Lsd (top rows) exhibit high-frequency variations, complex backgrounds, and structural artifacts. Minimizing the self-distillation distance (bottom rows) yields remarkably cleaner compositions. Meanwhile, the discriminator (right columns) enforces photorealistic textures and high fidelity, even on detailed images, and its score correlates with image quality, as noted in \cite{arjovsky_wasserstein_2017, azadi_discriminator_2019}.   


\subsection{Evolution of generated samples}

To quantify the stabilizing effect of the self-distillation objective, we measure the perceptual distance of the StyleGAN2-ADA generators across different training checkpoints (from $20\%$ to $100\%$ kimgs) using fixed latent codes. This metric functions similarly to the Perceptual Path Length \cite{karras_style-based_2019} but is  measured across the training trajectory, rather than through latent space interpolation. Table \ref{tab:sample_variance} reports the average perceptual distance and its standard deviation of the generated samples across these checkpoints, calculated using a sample size of 480. \mname significantly reduces the perceptual variance between samples compared to the baselines.

\begin{figure}
    \centering
    \begin{subfigure}[b]{\textwidth}
        \centering
        \offinterlineskip
        \includegraphics[width=\linewidth]{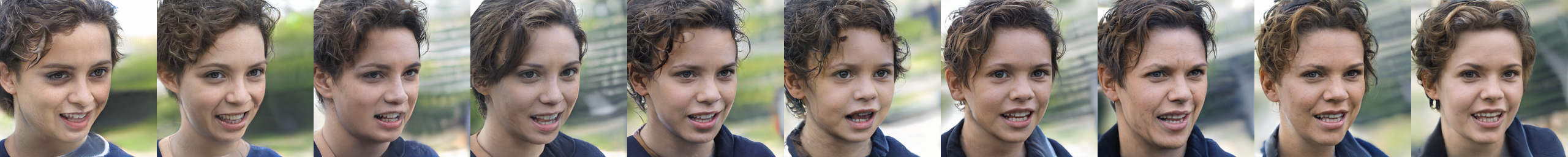}
        \includegraphics[width=\linewidth]{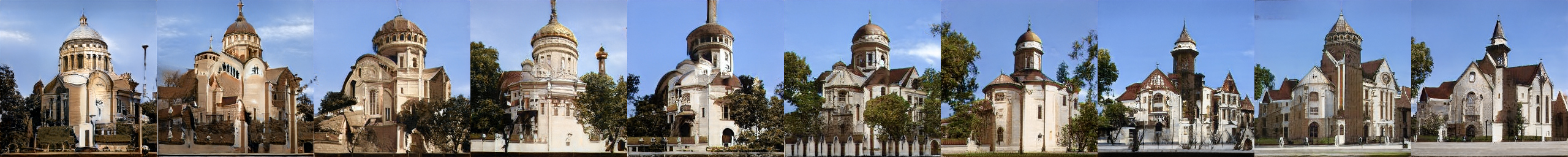}
    \end{subfigure}
    
    \vspace{0.3cm}
    
    \begin{subfigure}{\textwidth}
        \centering
        \offinterlineskip
        \includegraphics[width=\linewidth]{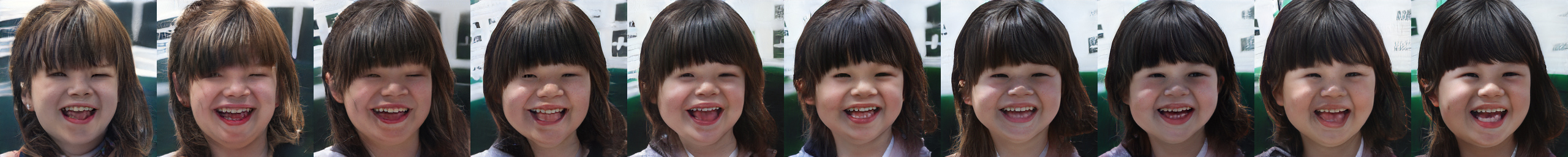}
        \includegraphics[width=\linewidth]{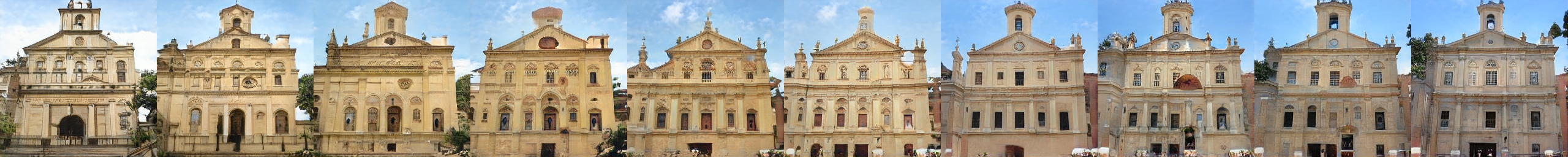}
    \end{subfigure}
    
    \caption{Evolution of generated images for a fixed latent vector $z$ during training. Top two rows: StyleGAN2-ADA; bottom two rows: StyleGAN2-ADA+SD. See Figure \ref{fig:big_evolution} for more. }
    \label{fig:evolution}
\end{figure}

To illustrate the visual effect of the above stabilization, in Fig.\ \ref{fig:evolution} we present the evolution of generated images for a fixed latent vector $z$ throughout the training process, for the sequences whose average perceptual distance is closest to the mean value reported in Table \ref{tab:sample_variance}. The images generated by the baseline model exhibit very high variance throughout the run; the facial features and church architecture change drastically. In contrast, the face generated by \mname for $z$ remains relatively stable, and could likely be identified as the same person in all shown images. Similarly, the overall structure of the church is quite consistent.

\begin{table}[b]
\centering
\caption{Average variation of generated samples across training checkpoints.}
\label{tab:sample_variance}
\begin{tabular}{lcccc}
\toprule
 & \multicolumn{2}{c}{FFHQ} & \multicolumn{2}{c}{LSUN Church} \\
\cmidrule(lr){2-3} \cmidrule(lr){4-5}
Model & Baseline & SD-GAN & Baseline & SD-GAN \\
\midrule
StyleGAN2-ADA & \tpm{0.375}{0.020} & \tpm{0.295}{0.028} & \tpm{0.482}{0.016} & \tpm{0.358}{0.033} \\
ProjectedGAN  & \tpm{0.470}{0.028} & \tpm{0.411}{0.022} & \tpm{0.513}{0.030} & \tpm{0.472}{0.029} \\
\bottomrule
\end{tabular}
\end{table}

\subsection{Limitations}





We identify the following limitations: (i) The LPIPS-based \Lsd may entrench ImageNet-induced biases and lead to ``feature leakage''. Though the results on the rFID metric in Table \ref{tab:FID_comparison} partially alleviate this concern, an even better mitigation of this bias is the choice of non-perceptual losses, which do not seem to lag far behind (as shown in ablations in Fig.\ \ref{fig:ablations}), or a discriminator-based \Lsd (Sec.\ \ref{sec:disc-based}). 
(ii) Our empirical evaluation covers four representative GAN architectures and three datasets; broader validation across additional GAN families, datasets, and resolutions remains future work. The proof in Sec.\ \ref{sec:method} is limited to the continuous-time, linearized Dirac-GAN with quadratic SD loss, and as such does not establish convergence for high-dimensional nonlinear GANs with stochastic optimization, perceptual losses, or differentiable augmentations.
(iii) Finally, \mname imposes relatively high computational overhead on training, especially when using perceptual \Lsd. Again, turning to non-perceptual SD losses may alleviate this problem. 


\section{Conclusions}

While nowadays it is diffusion models, rather than GANs, that are the primary tools for image generation, they are often computationally very demanding during inference. Distilling diffusion models into GANs has thus become a common technique to address this deficiency \cite{sauer_adversarial_2024, sauer_fast_2024, kang_distilling_2024}. This makes \mname a potentially useful tool for aiding this process, e.g. fine-tuning the GAN models distilled from diffusion models.


\begin{ack}
This research was funded by the Polish Ministry of Science and Higher Education, grant no. 0311/SBAD/0770, and by the National Science Centre, Poland, grant no. 2025/57/B/ST6/03737. For the purpose of Open Access, the authors have applied a CC BY public copyright license to any Author Accepted Manuscript (AAM) version arising from this submission.
\end{ack}

\printbibliography

\clearpage
\appendix

\section{Software implementation and deployment}\label{app:soft-impl}

The \mname variants used in this study have been implemented by minimally modifying the publicly available `official' implementations of the reference methods:
\begin{itemize}
    \item StyleGAN2-ADA: \url{https://github.com/nvlabs/stylegan2-ada-pytorch}
    \item ProjectedGAN: \url{https://github.com/autonomousvision/projected-gan}
    \item StyleNAT: \url{https://github.com/SHI-Labs/StyleNAT}
    \item StyleGAN-XL: \url{https://github.com/autonomousvision/stylegan-xl}
\end{itemize}

The infrastructure consisted of 1 Nvidia H100 GPU, 8 CPU cores, and 32 GB RAM for the single GPU setup and 4 $\times$ Nvidia H100 GPUs, 32 CPU cores and 128 GB RAM for 4 GPU training runs. The total estimated compute we used for all experiments is approximately $10\text{k}$ GPU-hours. The full specification of the nodes we used is:
\begin{itemize}
  \item CPU: AMD EPYC 9334, $2 \times 32$ cores
  \item RAM: 768 GB
  \item GPU: $4 \times$ Nvidia H100-94 SXM5, 94GB HBM2e
\end{itemize}

The implementation of our proposed method, including the training scripts and evaluation routines, is publicly available at \url{https://github.com/antollo/sd-gan}. Additionally, a supplementary project page featuring animated visualizations of our generated results is available at \url{https://antollo.github.io/sd-gan}.

Typical training runs of \mname using 4 GPUs on the $256^2$ datasets took respectively 46 h, 20 h, and 14 days for StyleGAN2-ADA+SD, ProjectedGAN+SD, and StyleNAT+SD. In comparison, their respective non-SD baselines took  41 h, 16 h, and 14 days. The baseline runs exhibited runtimes that are close to the \mname runs; this is because, during the baseline experiments, we still explicitly computed the difference between \G and \Gema (i.e., the \Lsd value) solely for tracking and monitoring purposes, without using it to update the network.

\section{Challenges in replication of prior results}\label{app:replication}

Due to the hardware constraints of the Nvidia H100 GPUs (Hopper architecture, \texttt{sm\_90}) utilized in our experiments, we were required to upgrade PyTorch to ensure compatibility for both StyleGAN2-ADA and ProjectedGAN (compared to the versions used in the original implementations). Specifically, our environment used \texttt{torch==2.1.2} compiled with CUDA 12.1.

After several unsuccessful attempts to match the baseline results for StyleNAT on LSUN Church, we excluded this combination from our study. We contacted the authors via a GitHub issue\footnote{\url{https://github.com/SHI-Labs/StyleNAT/issues/18}},
and while they were very helpful and provided feedback, we ultimately failed to achieve outcomes consistent with the reported benchmarks. It is worth noting that the Church configuration file is missing from their repository, leading us to believe that an omitted or forgotten technical detail may be responsible for the discrepancy.

We also encountered replication challenges with ProjectedGAN. The original work utilized an 8-GPU training setup. In the official implementation in a distributed multi-device setting, Batch Normalization layers compute statistics separately on each device's mini-batch fragment. This inadvertently applies the concept of \emph{Ghost Batch Normalization} \cite{hoffer_train_2017}, which is known to aid generalization and yield better results than computing statistics over the entire batch. To simulate this 8-GPU dynamic on our 4-GPU setup, we explicitly implemented Ghost BN. However, even with this correction, the best baseline FID we achieved on LSUN Church was 2.11, falling short of the 1.59 reported in the original paper. Furthermore, evaluating the authors' officially provided model (weights) across various repository commits also failed to reproduce the reported FID. This replication issue appears to be a known problem, mentioned in other independent studies \cite{stein_exposing_2023} and GitHub issues\footnote{\url{https://github.com/autonomousvision/projected-gan/issues/103}}.

Regarding StyleGAN-XL, we attempted to replicate the results on ImageNet at $64^2$ resolution. The original study did not use a fixed training schedule; rather than that, the authors noted that each progressive growth stage was trained until the FID stopped decreasing \cite[Sec.~3.2]{sauer_stylegan-xl_2022}. They reported a final FID of 1.51 at this resolution. In our experiments, we initialized the baseline and SD variant from the same pretrained $32^2$ stem and trained each of them for $10k$ kimgs at $64^2$. The baseline run required approximately 1.3 days on H100 GPU and achieved FID of 2.24. While the original work did not provide an exact training duration for the $64^2$ stage, the authors noted that the subsequent $128^2$ resolution required 15 V100-days, which may suggest that their training budget was significantly larger than ours.

Beyond the specific cases of StyleNAT and ProjectedGAN, we encountered similar replication failures with several other open-source repositories. While we optimistically assume that these discrepancies stem from our own experimental misconfigurations, these repeated challenges serve as a reminder that reproducibility remains a significant issue.

\section{Pseudocode}\label{app:pseudocode}

Listing \ref{lst:attention_pseudo} presents the pseudocode of \mname, as implemented in PyTorch. 

\begin{listing}[htbp]
\caption{PyTorch pseudocode for the \mname generator update step. The highlighted section demonstrates the calculation of the self-distillation loss.}
\label{lst:attention_pseudo}
\begin{minted}[
    bgcolor=codebg,
    baselinestretch=1.1,
    fontsize=\footnotesize,
    linenos,
    xleftmargin=2em,
    highlightlines={18,19,20,21,22,23,24,25,26},
    highlightcolor=focusbg
]{python}
loss_type = 'LPIPS' # or 'L1'
use_aug = True # or False
aug2 = AugmentationSequential(
  RandomHorizontalFlip(p=0.5),
  RandomRotate(degrees=5, p=0.5),
  RandomTranslate(translate=(0.01, 0.01), p=0.5),
  random_apply=2
)

# Inside the training loop:
z = torch.randn([batch_size, z_dim])
fake_images = G(z)
fake_score = D(fake_images)
# Normal adversarial loss, e.g.
adv_loss = -torch.mean(fake_score)
G_loss = adv_loss.mean()
# Self-distillation loss
with torch.no_grad():
    fake_images_ema = G_ema(z)
if use_aug:
    fake_images, fake_images_ema = aug2(fake_images, fake_images_ema)
if loss_type == 'L1':
    SD_loss = l1_loss(fake_images, fake_images_ema)
elif loss_type == 'LPIPS':
    SD_loss = lpips(fake_images, fake_images_ema, net='vgg')
G_loss = G_loss + alpha * SD_loss.mean()
G_loss.backward()
# Continue the training loop
\end{minted}
\end{listing}

\section{Comparison with alternative temporal regularization baselines}\label{app:temporal-baselines}

To isolate the performance gains of \mname that originate specifically from our combination of output-space consistency and EMA, we compared it against a simpler temporal regularization baseline: a delayed generator checkpoint with weight-space regularization. Specifically, we maintained a queue of $N=10$ historical model weights, updated every $T/N$ kimgs (setting $T=20$ to match the EMA half-life from our main experiments; see Table~\ref{tab:FID_comparison}). We then applied an $L_2$ penalty between the weights of the active generator and the oldest historical weights from the queue. Table \ref{tab:weight_space_ablation} reports the FID scores for ProjectedGAN on FFHQ at $256^2$ across various penalty weights $\lambda$.

\begin{table}
\centering
\caption{Comparison of the final FID scores on FFHQ at $256^2$ achieved by the baseline ProjectedGAN, the variant with the weight-space temporal regularization for different penalty weights $\lambda$, and \mname.
}\label{tab:weight_space_ablation}
\begin{tabular}{lll}
\toprule
\multicolumn{2}{l}{Model}& FID\down \\
\midrule
\multicolumn{2}{l}{ProjectedGAN}            & 3.90   \\
\midrule
\multirow{6}{*}{\begin{tabular}[c]{@{}l@{}}ProjectedGAN\\ (w/ t. reg.)\end{tabular}} 
                         & $\lambda = 10^5$ & 4.70   \\
                         & $\lambda = 10^4$ & 4.24   \\
                         & $\lambda = 10^3$ & 3.85   \\
                         & $\lambda = 10^2$ & 3.96   \\
                         & $\lambda = 10$   & 3.96   \\
                         & $\lambda = 1$    & 4.20   \\
\midrule
\multicolumn{2}{l}{ProjectedGAN+SD}         & 3.15  \\
\bottomrule
\end{tabular}
\end{table}

\begin{figure}
  \centering\includegraphics[trim={7pt 7pt 7pt 7pt}, clip, width=0.85\linewidth]{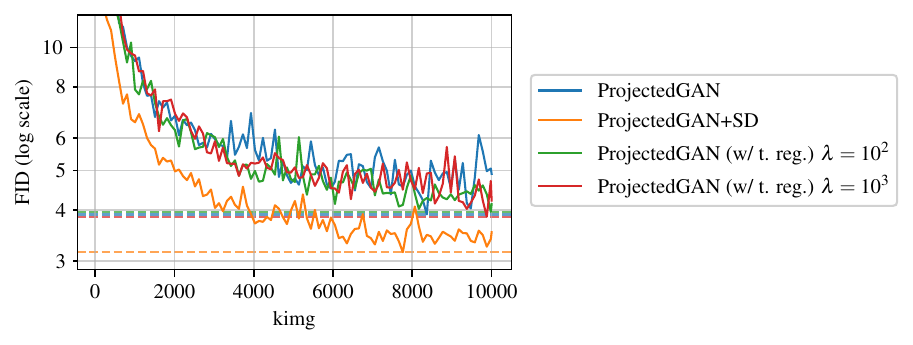}
  \caption{FID scores (log scale) during training on the FFHQ dataset at $256^2$ comparing the baseline ProjectedGAN, \mname, and the best weight-space temporal regularization variants. Horizontal dashed lines indicate the best FID score achieved by the respective models.
  }\label{fig:weight_space_ablation}
\end{figure}

While this temporal smoothing approach yields a negligible improvement over the unregularized baseline at one of the tested penalty weights ($\lambda = 10^3$), yielding an FID of 3.85, it falls significantly short of \mname (FID 3.15). Furthermore, the learning curves in Figure \ref{fig:weight_space_ablation} demonstrate that while temporal regularization dampens variance compared to the baseline, it fails to lead to deeper convergence achieved by self-distillation. This confirms that simple weight-space temporal anchoring does not replicate the gains of our method, underscoring the importance of the output-space consistency combined with EMA.

\section{Auxiliary results}\label{app:aux-results}

\begin{figure}[!ht]
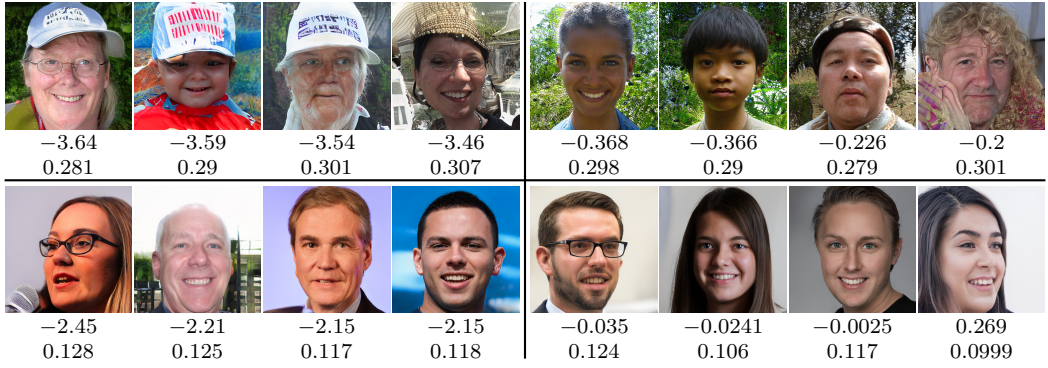
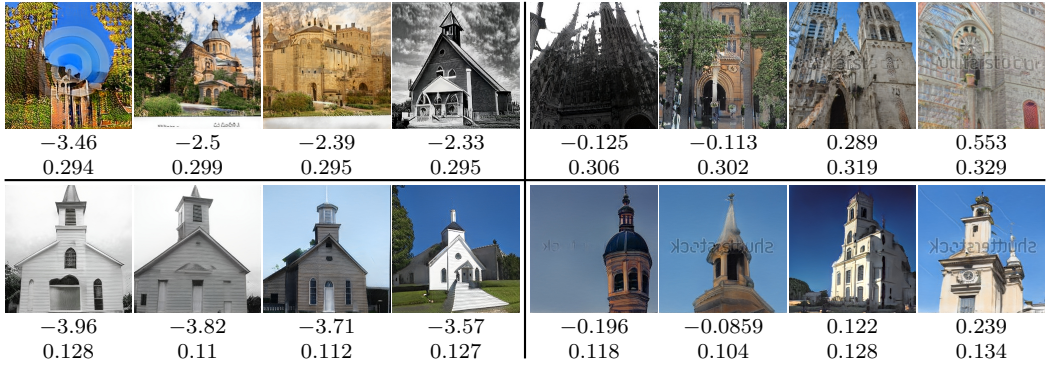

    \centering
    \begin{subfigure}[b]{\textwidth}
    \centering
    \begin{tikzpicture}[
        node distance = 0.5pt,
        every node/.style = {inner sep=0pt}
    ]
    
      \node (tl1) {\includegraphics[width=0.12\linewidth]{figures/a/stylegan2-ada-ffhq256/network-snapshot-023889_dissimilar_fake_SD0.2809_D-3.6358.jpg}};
      \node (tl1-lbl) [below=2pt of tl1, font=\footnotesize, align=center] {\rl{0.2809}{-3.6358}};

      \node (tl2) [right=of tl1] {\includegraphics[width=0.12\linewidth]{figures/a/stylegan2-ada-ffhq256/network-snapshot-023889_dissimilar_fake_SD0.2904_D-3.5898.jpg}};
      \node [below=2pt of tl2, font=\footnotesize, align=center] {\rl{0.2904}{-3.5898}};

      \node (tl3) [right=of tl2] {\includegraphics[width=0.12\linewidth]{figures/a/stylegan2-ada-ffhq256/network-snapshot-023889_dissimilar_fake_SD0.3006_D-3.5395.jpg}};
      \node [below=2pt of tl3, font=\footnotesize, align=center] {\rl{0.3006}{-3.5395}};

      \node (tl4) [right=of tl3] {\includegraphics[width=0.12\linewidth]{figures/a/stylegan2-ada-ffhq256/network-snapshot-023889_dissimilar_fake_SD0.3068_D-3.4634.jpg}};
      \node [below=2pt of tl4, font=\footnotesize, align=center] {\rl{0.3068}{-3.4634}};

      \node (tr1) [right=4pt of tl4] {\includegraphics[width=0.12\linewidth]{figures/a/stylegan2-ada-ffhq256/network-snapshot-023889_dissimilar_real_SD0.2980_D-0.3675.jpg}};
      \node [below=2pt of tr1, font=\footnotesize, align=center] {\rl{0.2980}{-0.3675}};

      \node (tr2) [right=of tr1] {\includegraphics[width=0.12\linewidth]{figures/a/stylegan2-ada-ffhq256/network-snapshot-023889_dissimilar_real_SD0.2898_D-0.3657.jpg}};
      \node [below=2pt of tr2, font=\footnotesize, align=center] {\rl{0.2898}{-0.3657}};

      \node (tr3) [right=of tr2] {\includegraphics[width=0.12\linewidth]{figures/a/stylegan2-ada-ffhq256/network-snapshot-023889_dissimilar_real_SD0.2785_D-0.2261.jpg}};
      \node [below=2pt of tr3, font=\footnotesize, align=center] {\rl{0.2785}{-0.2261}};

      \node (tr4) [right=of tr3] {\includegraphics[width=0.12\linewidth]{figures/a/stylegan2-ada-ffhq256/network-snapshot-023889_dissimilar_real_SD0.3005_D-0.2001.jpg}};
      \node [below=2pt of tr4, font=\footnotesize, align=center] {\rl{0.3005}{-0.2001}};

      \node (bl1) [below=4pt of tl1-lbl] {\includegraphics[width=0.12\linewidth]{figures/a/stylegan2-ada-ffhq256/network-snapshot-023889_similar_fake_SD0.1280_D-2.4491.jpg}};
      \node (bl1-lbl) [below=2pt of bl1, font=\footnotesize, align=center] {\rl{0.1280}{-2.4491}};

      \node (bl2) [right=of bl1] {\includegraphics[width=0.12\linewidth]{figures/a/stylegan2-ada-ffhq256/network-snapshot-023889_similar_fake_SD0.1250_D-2.2094.jpg}};
      \node [below=2pt of bl2, font=\footnotesize, align=center] {\rl{0.1250}{-2.2094}};

      \node (bl3) [right=of bl2] {\includegraphics[width=0.12\linewidth]{figures/a/stylegan2-ada-ffhq256/network-snapshot-023889_similar_fake_SD0.1171_D-2.1547.jpg}};
      \node [below=2pt of bl3, font=\footnotesize, align=center] {\rl{0.1171}{-2.1547}};

      \node (bl4) [right=of bl3] {\includegraphics[width=0.12\linewidth]{figures/a/stylegan2-ada-ffhq256/network-snapshot-023889_similar_fake_SD0.1178_D-2.1545.jpg}};
      \node [below=2pt of bl4, font=\footnotesize, align=center] {\rl{0.1178}{-2.1545}};

      \node (br1) [right=4pt of bl4] {\includegraphics[width=0.12\linewidth]{figures/a/stylegan2-ada-ffhq256/network-snapshot-023889_similar_real_SD0.1240_D-0.0350.jpg}};
      \node [below=2pt of br1, font=\footnotesize, align=center] {\rl{0.1240}{-0.0350}};

      \node (br2) [right=of br1] {\includegraphics[width=0.12\linewidth]{figures/a/stylegan2-ada-ffhq256/network-snapshot-023889_similar_real_SD0.1055_D-0.0241.jpg}};
      \node [below=2pt of br2, font=\footnotesize, align=center] {\rl{0.1055}{-0.0241}};

      \node (br3) [right=of br2] {\includegraphics[width=0.12\linewidth]{figures/a/stylegan2-ada-ffhq256/network-snapshot-023889_similar_real_SD0.1169_D-0.0025.jpg}};
      \node [below=2pt of br3, font=\footnotesize, align=center] {\rl{0.1169}{-0.0025}};

      \node (br4) [right=of br3] {\includegraphics[width=0.12\linewidth]{figures/a/stylegan2-ada-ffhq256/network-snapshot-023889_similar_real_SD0.0999_D0.2691.jpg}};
      \node (br4-lbl) [below=2pt of br4, font=\footnotesize, align=center] {\rl{0.0999}{0.2691}};

      \path (tl4.east) -- (tr1.west) coordinate[midway] (x-mid);
      \path (tl1-lbl.south) -- (bl1.north) coordinate[midway] (y-mid);
      \draw[-, thick] (tl1.west |- y-mid) -- (tr4.east |- y-mid);
      \draw[-, thick] (x-mid |- bl1-lbl.south) -- (x-mid |- tl1.north);

    \end{tikzpicture}
    \caption{StyleGAN2-ADA+SD, FFHQ at $256^2$.}
    \vspace*{2mm}
    \end{subfigure}

    \begin{subfigure}[b]{\textwidth}
    \centering
    \begin{tikzpicture}[
        node distance = 0.5pt,
        every node/.style = {inner sep=0pt}
    ]
    
      \node (tl1) {\includegraphics[width=0.12\linewidth]{figures/a/stylegan2-ada-church256/network-snapshot-024897_dissimilar_fake_SD0.2937_D-3.4573.jpg}};
      \node (tl1-lbl) [below=2pt of tl1, font=\footnotesize, align=center] {\rl{0.2937}{-3.4573}};

      \node (tl2) [right=of tl1] {\includegraphics[width=0.12\linewidth]{figures/a/stylegan2-ada-church256/network-snapshot-024897_dissimilar_fake_SD0.2988_D-2.5040.jpg}};
      \node [below=2pt of tl2, font=\footnotesize, align=center] {\rl{0.2988}{-2.5040}};

      \node (tl3) [right=of tl2] {\includegraphics[width=0.12\linewidth]{figures/a/stylegan2-ada-church256/network-snapshot-024897_dissimilar_fake_SD0.2947_D-2.3944.jpg}};
      \node [below=2pt of tl3, font=\footnotesize, align=center] {\rl{0.2947}{-2.3944}};

      \node (tl4) [right=of tl3] {\includegraphics[width=0.12\linewidth]{figures/a/stylegan2-ada-church256/network-snapshot-024897_dissimilar_fake_SD0.2953_D-2.3331.jpg}};
      \node [below=2pt of tl4, font=\footnotesize, align=center] {\rl{0.2953}{-2.3331}};

      \node (tr1) [right=4pt of tl4] {\includegraphics[width=0.12\linewidth]{figures/a/stylegan2-ada-church256/network-snapshot-024897_dissimilar_real_SD0.3055_D-0.1246.jpg}};
      \node [below=2pt of tr1, font=\footnotesize, align=center] {\rl{0.3055}{-0.1246}};

      \node (tr2) [right=of tr1] {\includegraphics[width=0.12\linewidth]{figures/a/stylegan2-ada-church256/network-snapshot-024897_dissimilar_real_SD0.3021_D-0.1133.jpg}};
      \node [below=2pt of tr2, font=\footnotesize, align=center] {\rl{0.3021}{-0.1133}};

      \node (tr3) [right=of tr2] {\includegraphics[width=0.12\linewidth]{figures/a/stylegan2-ada-church256/network-snapshot-024897_dissimilar_real_SD0.3194_D0.2890.jpg}};
      \node [below=2pt of tr3, font=\footnotesize, align=center] {\rl{0.3194}{0.2890}};

      \node (tr4) [right=of tr3] {\includegraphics[width=0.12\linewidth]{figures/a/stylegan2-ada-church256/network-snapshot-024897_dissimilar_real_SD0.3287_D0.5528.jpg}};
      \node [below=2pt of tr4, font=\footnotesize, align=center] {\rl{0.3287}{0.5528}};

      \node (bl1) [below=4pt of tl1-lbl] {\includegraphics[width=0.12\linewidth]{figures/a/stylegan2-ada-church256/network-snapshot-024897_similar_fake_SD0.1282_D-3.9614.jpg}};
      \node (bl1-lbl) [below=2pt of bl1, font=\footnotesize, align=center] {\rl{0.1282}{-3.9614}};

      \node (bl2) [right=of bl1] {\includegraphics[width=0.12\linewidth]{figures/a/stylegan2-ada-church256/network-snapshot-024897_similar_fake_SD0.1096_D-3.8214.jpg}};
      \node [below=2pt of bl2, font=\footnotesize, align=center] {\rl{0.1096}{-3.8214}};

      \node (bl3) [right=of bl2] {\includegraphics[width=0.12\linewidth]{figures/a/stylegan2-ada-church256/network-snapshot-024897_similar_fake_SD0.1122_D-3.7081.jpg}};
      \node [below=2pt of bl3, font=\footnotesize, align=center] {\rl{0.1122}{-3.7081}};

      \node (bl4) [right=of bl3] {\includegraphics[width=0.12\linewidth]{figures/a/stylegan2-ada-church256/network-snapshot-024897_similar_fake_SD0.1268_D-3.5707.jpg}};
      \node [below=2pt of bl4, font=\footnotesize, align=center] {\rl{0.1268}{-3.5707}};

      \node (br1) [right=4pt of bl4] {\includegraphics[width=0.12\linewidth]{figures/a/stylegan2-ada-church256/network-snapshot-024897_similar_real_SD0.1178_D-0.1959.jpg}};
      \node [below=2pt of br1, font=\footnotesize, align=center] {\rl{0.1178}{-0.1959}};

      \node (br2) [right=of br1] {\includegraphics[width=0.12\linewidth]{figures/a/stylegan2-ada-church256/network-snapshot-024897_similar_real_SD0.1035_D-0.0859.jpg}};
      \node [below=2pt of br2, font=\footnotesize, align=center] {\rl{0.1035}{-0.0859}};

      \node (br3) [right=of br2] {\includegraphics[width=0.12\linewidth]{figures/a/stylegan2-ada-church256/network-snapshot-024897_similar_real_SD0.1283_D0.1224.jpg}};
      \node [below=2pt of br3, font=\footnotesize, align=center] {\rl{0.1283}{0.1224}};

      \node (br4) [right=of br3] {\includegraphics[width=0.12\linewidth]{figures/a/stylegan2-ada-church256/network-snapshot-024897_similar_real_SD0.1338_D0.2387.jpg}};
      \node (br4-lbl) [below=2pt of br4, font=\footnotesize, align=center] {\rl{0.1338}{0.2387}};

      \path (tl4.east) -- (tr1.west) coordinate[midway] (x-mid);
      \path (tl1-lbl.south) -- (bl1.north) coordinate[midway] (y-mid);
      \draw[-, thick] (tl1.west |- y-mid) -- (tr4.east |- y-mid);
      \draw[-, thick] (x-mid |- bl1-lbl.south) -- (x-mid |- tl1.north);
      
    \end{tikzpicture}
    \caption{StyleGAN2-ADA+SD, LSUN Church at $256^2$.}
    \end{subfigure}
    
    \caption{Qualitative interplay between the objectives (detailed version of Fig.\ \ref{fig:sd_vs_adv}). \textbf{Left vs. right:} smaller vs. larger \textit{realness} according to \D (\D logits, top value). \textbf{Top vs. bottom:} larger vs. smaller difference between \G and \Gema (\Lsd, bottom value under image). The adversarial objective pulls generator to the right, while the self-distillation objective pulls it to the bottom. Best viewed when zoomed in.}
    \label{fig:sd_vs_adv_detailed}
\end{figure}

\begin{figure}[!ht]
    \centering
    
    \begin{subfigure}[b]{\textwidth}
        \centering
        \includegraphics[width=\linewidth]{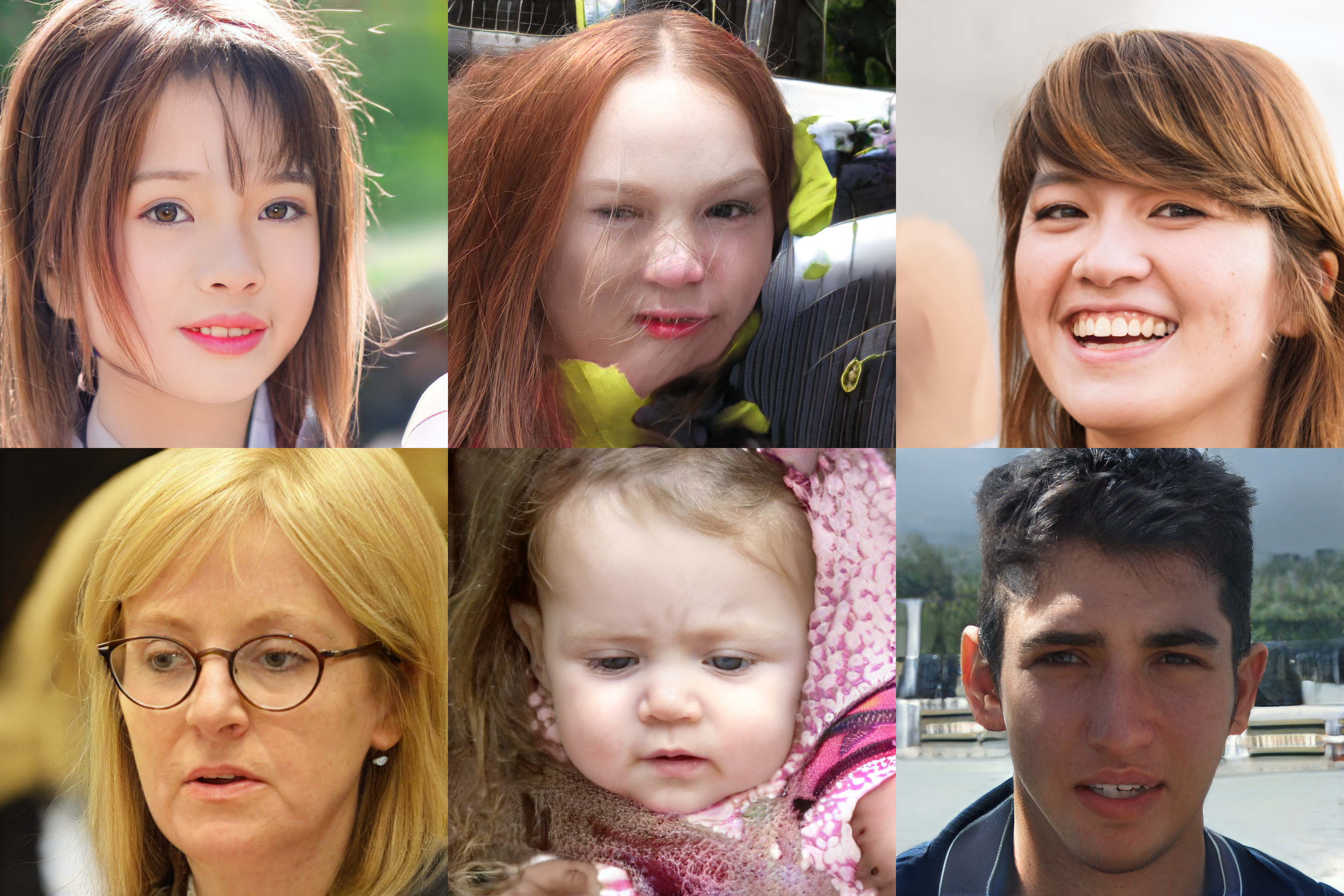}
        \caption{StyleGAN2-ADA (FFHQ $1024^2$ fine-tuning)}
    \end{subfigure}
    
    \vspace{0.3cm}
    
    \begin{subfigure}[b]{\textwidth}
        \centering
        \includegraphics[width=\linewidth]{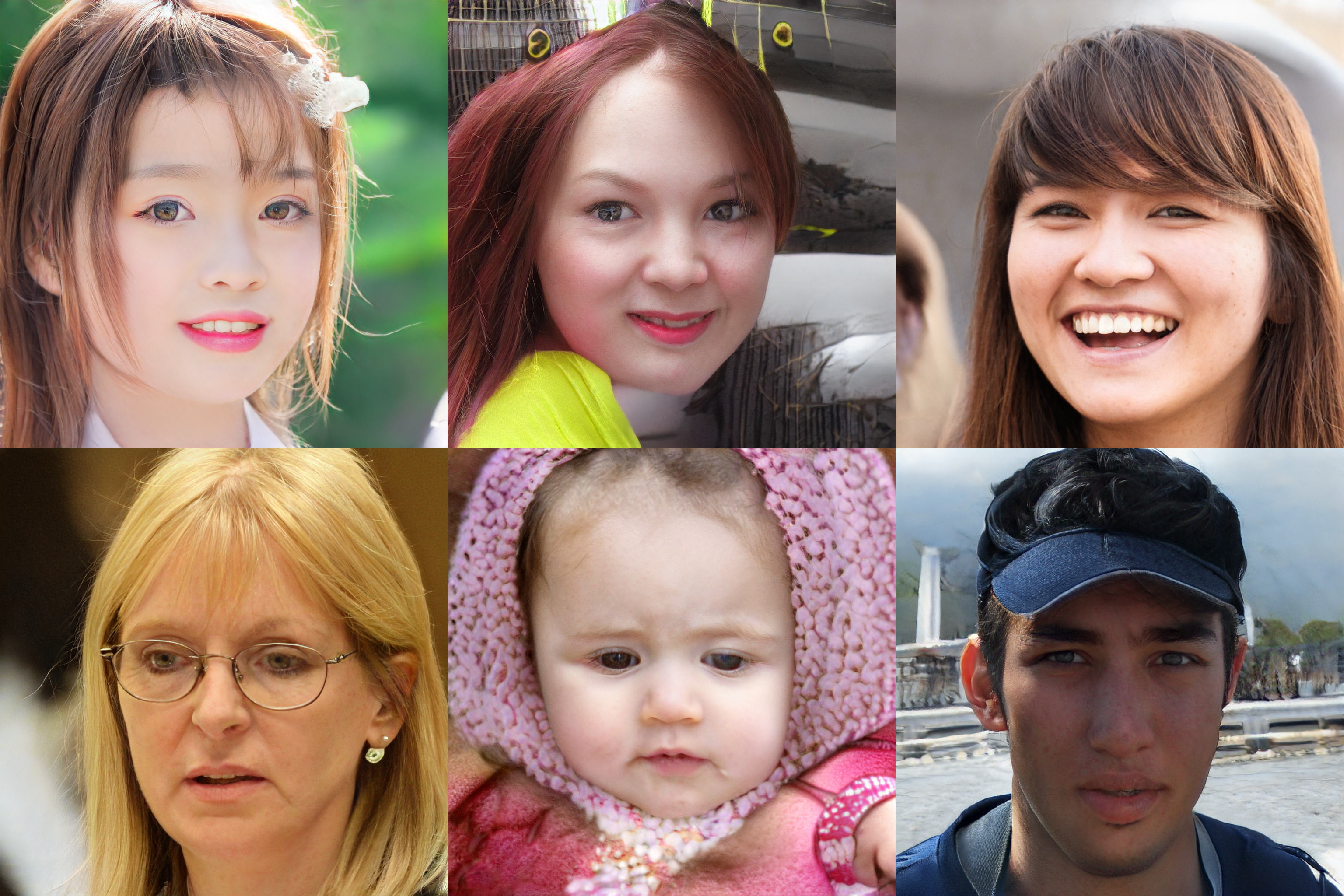}
        \caption{StyleGAN2-ADA+SD (FFHQ $1024^2$ fine-tuning)}
    \end{subfigure}
    
    \caption{Qualitative results after fine-tuning on FFHQ at $1024^2$. Samples generated using the same latent vectors for both methods are shown for comparison.}
    \label{fig:finetune_1024}
\end{figure}

Figure \ref{fig:sd_vs_adv_detailed} presents the detailed version of Figure \ref{fig:sd_vs_adv}, with additional information on both objectives. The evolution of images generated during training for a fixed latent vector $z$ is shown in Figure \ref{fig:big_evolution}. Figure \ref{fig:finetune_1024} provides a qualitative comparison of the baseline and \mname after fine-tuning on the FFHQ dataset at $1024^2$ resolution.

\begin{figure}[!ht]
    \centering
    
    \begin{subfigure}[b]{\textwidth}
        \centering
        \offinterlineskip
        \includegraphics[width=\linewidth]{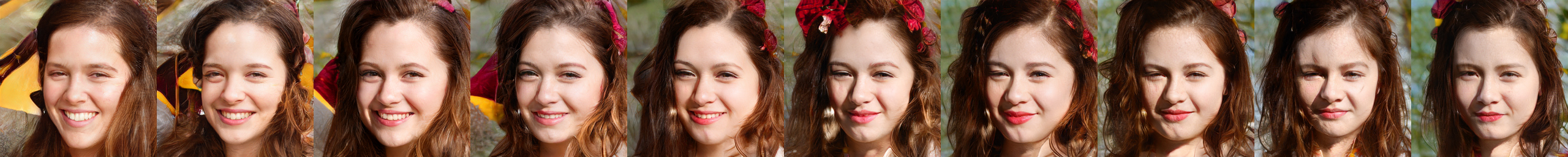}
        \includegraphics[width=\linewidth]{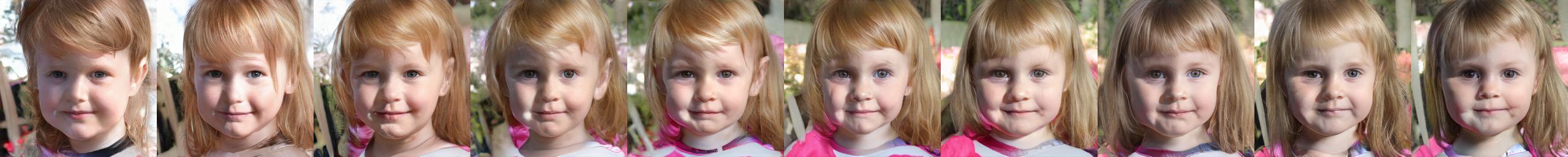}
        \includegraphics[width=\linewidth]{figures/stitches/style_ffhq_sd0/stitch_patch_293_progression.jpg}
        \caption{StyleGAN2-ADA (FFHQ)}
    \end{subfigure}
    
    \vspace{0.3cm}
    
    \begin{subfigure}[b]{\textwidth}
        \centering
        \offinterlineskip
        \includegraphics[width=\linewidth]{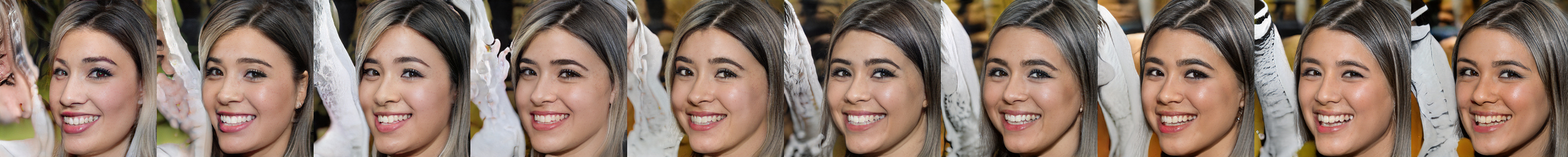}
        \includegraphics[width=\linewidth]{figures/stitches/style_ffhq_sd1/stitch_patch_355_progression.jpg}
        \includegraphics[width=\linewidth]{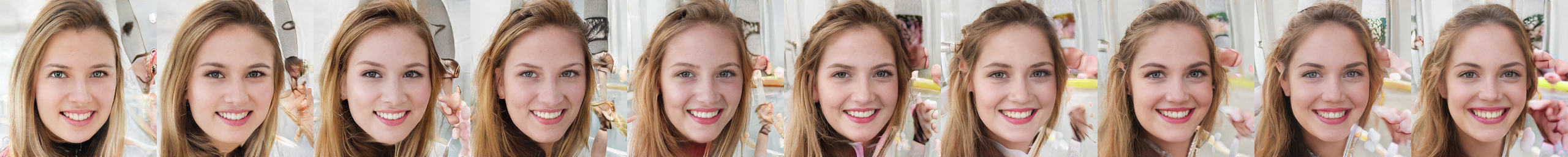}
        \caption{StyleGAN2-ADA+SD (FFHQ)}
    \end{subfigure}
    
    \vspace{0.3cm}
    
    \begin{subfigure}[b]{\textwidth}
        \centering
        \offinterlineskip
        \includegraphics[width=\linewidth]{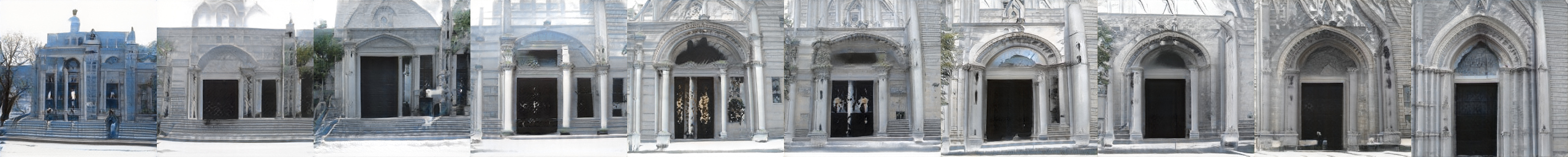}
        \includegraphics[width=\linewidth]{figures/stitches/style_church_sd0/stitch_patch_414_progression.jpg}
        \includegraphics[width=\linewidth]{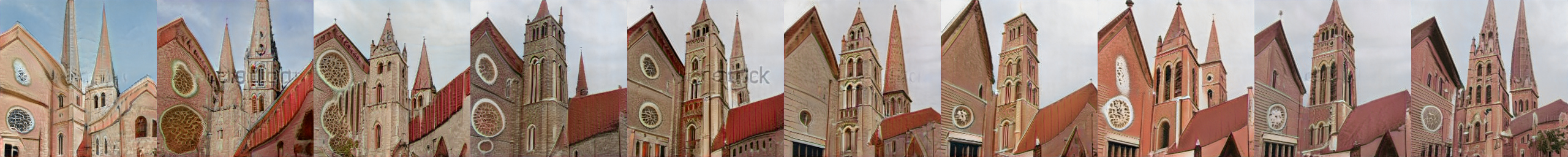}
        \caption{StyleGAN2-ADA (LSUN Church)}
    \end{subfigure}
    
    \vspace{0.3cm}
    
    \begin{subfigure}[b]{\textwidth}
        \centering
        \offinterlineskip
        \includegraphics[width=\linewidth]{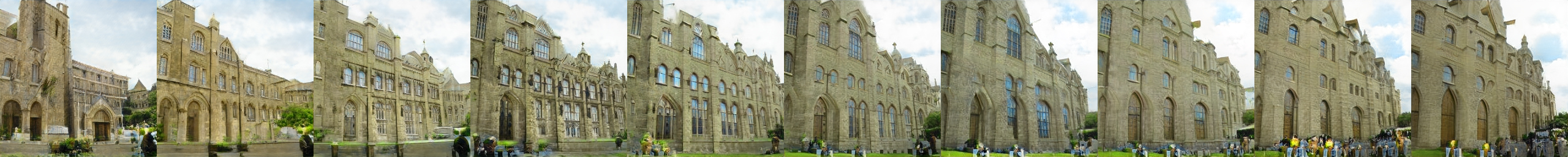}
        \includegraphics[width=\linewidth]{figures/stitches/style_church_sd1/stitch_patch_436_progression.jpg}
        \includegraphics[width=\linewidth]{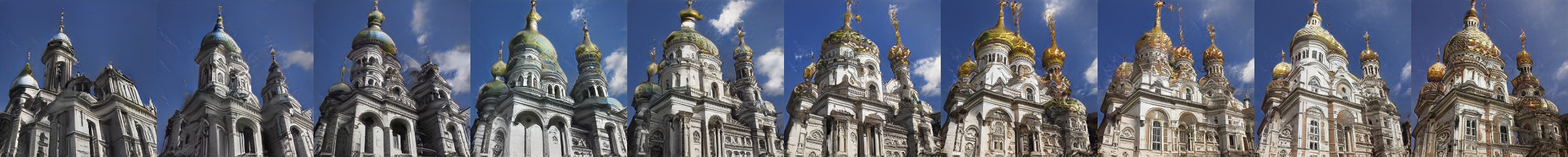}
        \caption{StyleGAN2-ADA+SD (LSUN Church)}
    \end{subfigure}
    
    \caption{Evolution of images generated during training for a fixed latent vector $z$. \textit{(Continued on next page)}}
\end{figure}

\begin{figure}[!ht]
    \ContinuedFloat
    \centering
    
    \begin{subfigure}[b]{\textwidth}
        \centering
        \offinterlineskip
        \includegraphics[width=\linewidth]{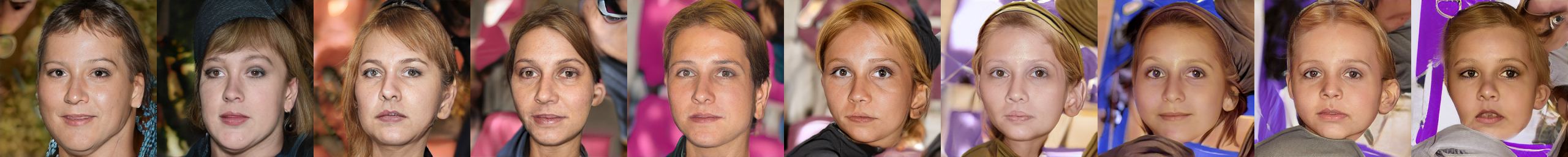}
        \includegraphics[width=\linewidth]{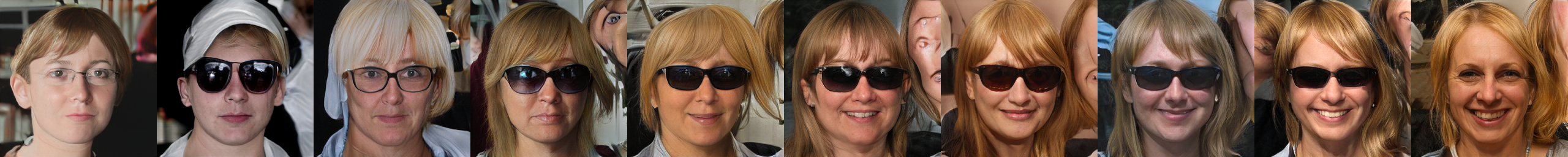}
        \includegraphics[width=\linewidth]{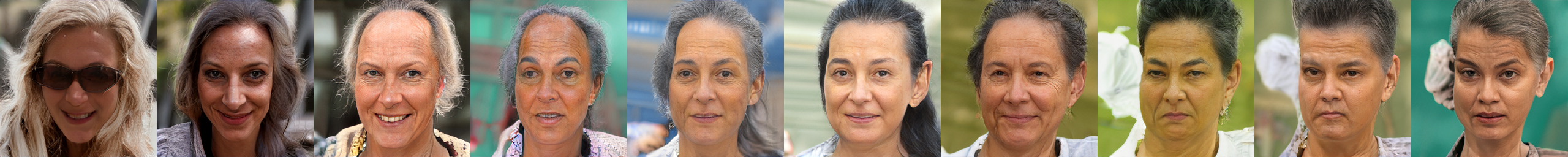}
        \caption{ProjectedGAN (FFHQ)}
    \end{subfigure}
    
    \vspace{0.3cm}
    
    \begin{subfigure}[b]{\textwidth}
        \centering
        \offinterlineskip
        \includegraphics[width=\linewidth]{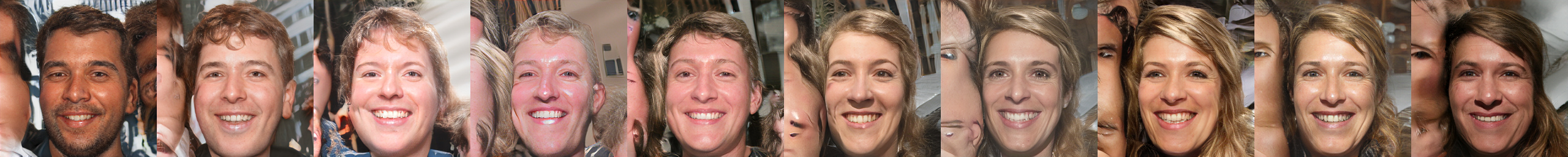}
        \includegraphics[width=\linewidth]{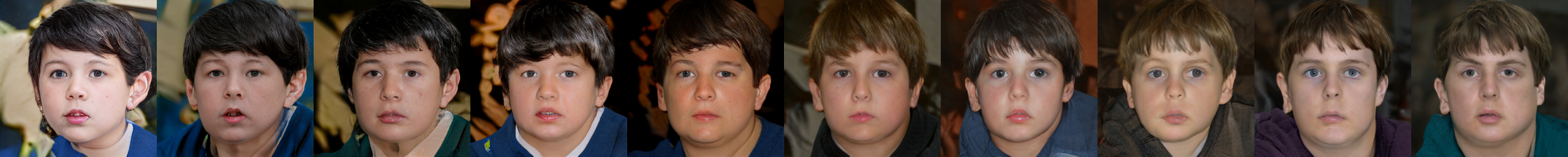}
        \includegraphics[width=\linewidth]{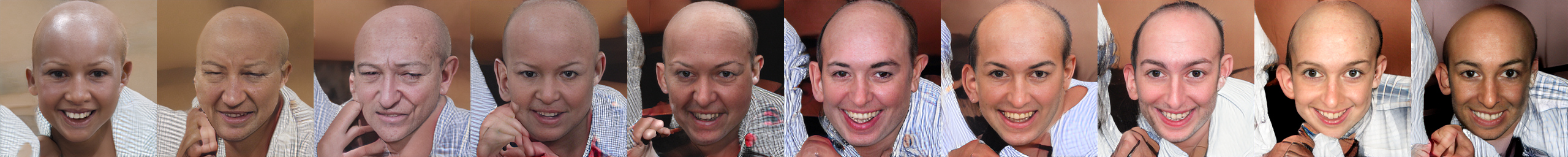}
        \caption{ProjectedGAN+SD (FFHQ)}
        \label{fig:evolution_adasd_ffhq}
    \end{subfigure}
    
    \vspace{0.3cm}
    
    \begin{subfigure}[b]{\textwidth}
        \centering
        \offinterlineskip
        \includegraphics[width=\linewidth]{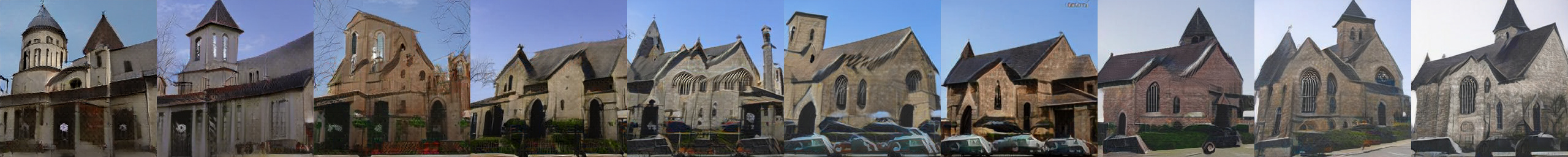}
        \includegraphics[width=\linewidth]{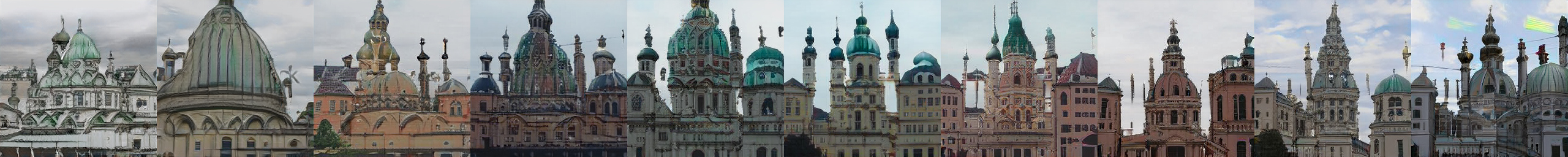}
        \includegraphics[width=\linewidth]{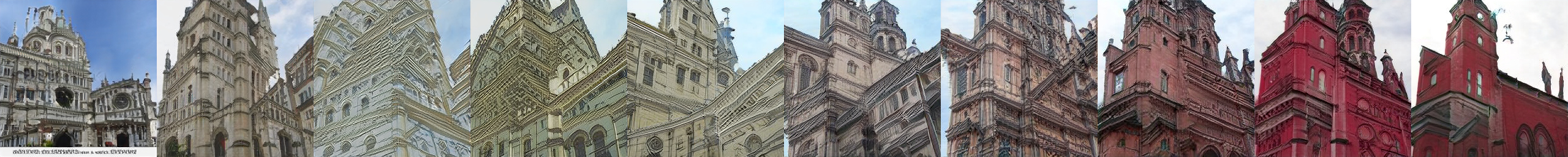}
        \caption{ProjectedGAN (LSUN Church)}
        \label{fig:evolution_ada_church}
    \end{subfigure}
    
    \vspace{0.3cm}
    
    \begin{subfigure}[b]{\textwidth}
        \centering
        \offinterlineskip
        \includegraphics[width=\linewidth]{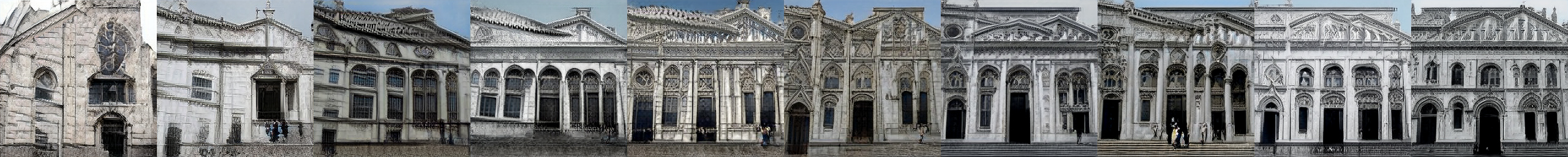}
        \includegraphics[width=\linewidth]{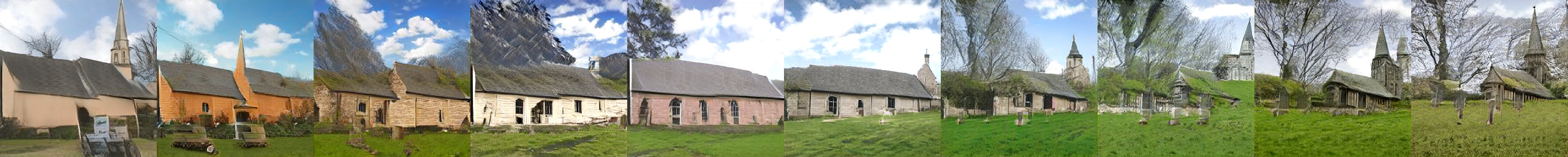}
        \includegraphics[width=\linewidth]{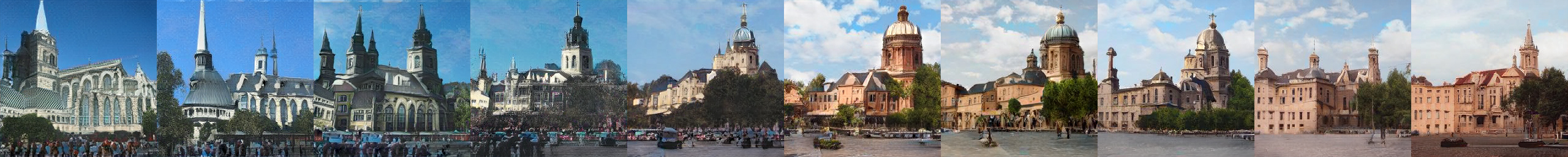}
        \caption{ProjectedGAN+SD (LSUN Church)}
        \label{fig:evolution_adasd_church}
    \end{subfigure}

    \caption{Evolution of images generated during training for a fixed latent vector $z$. \textit{(Continued on next page)}}
\end{figure}

\begin{figure}[!ht]
    \ContinuedFloat
    \centering
    
    \begin{subfigure}[b]{\textwidth}
        \centering
        \offinterlineskip
        \includegraphics[width=\linewidth]{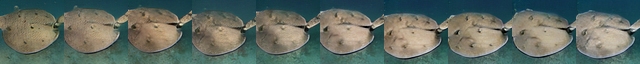}
        \includegraphics[width=\linewidth]{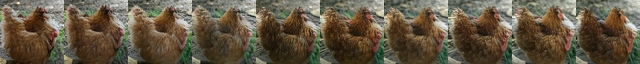}
        \includegraphics[width=\linewidth]{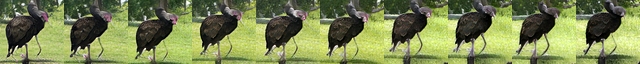}
        \includegraphics[width=\linewidth]{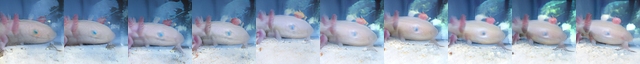}
        \caption{StyleGAN-XL (ImageNet at $64^2$, classes: 5 electric ray, 8 hen, 23 vulture, and 29 axolotl)}
    \end{subfigure}

    \vspace{0.3cm}

    \begin{subfigure}[b]{\textwidth}
        \centering
        \offinterlineskip
        \includegraphics[width=\linewidth]{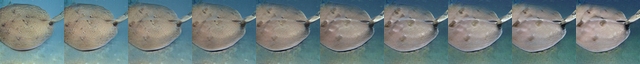}
        \includegraphics[width=\linewidth]{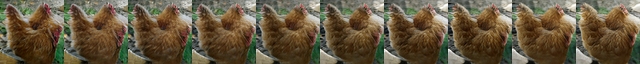}
        \includegraphics[width=\linewidth]{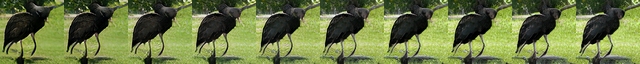}
        \includegraphics[width=\linewidth]{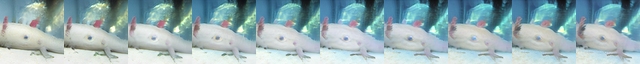}
        \caption{StyleGAN-XL+SD (ImageNet at $64^2$, classes: 5 electric ray, 8 hen, 23 vulture, and 29 axolotl)}
    \end{subfigure}
    
    \vspace{0.3cm}
    
    \caption{Evolution of images generated during training for a fixed latent vector $z$. The progressions are grouped by model and dataset (detailed version of Figure \ref{fig:evolution}).}
    \label{fig:big_evolution}
\end{figure}

\end{document}